\documentclass[twoside]{article}

\usepackage[accepted]{aistats2020}
\usepackage[round]{natbib}

\usepackage[utf8]{inputenc} 
\usepackage[T1]{fontenc}    
\usepackage{hyperref}       
\usepackage{url}            
\usepackage{booktabs}       
\usepackage{amsfonts}       
\usepackage{nicefrac}       
\usepackage{microtype}      
\usepackage{amsmath}
\usepackage{mathtools}
\usepackage{amssymb}
\usepackage{amsthm}
\usepackage[abs]{overpic}
\usepackage[usenames]{xcolor}
\usepackage{graphicx}
\usepackage[labelfont=bf]{caption}
\usepackage{subcaption}
\usepackage{float}
\usepackage{dsfont}
\usepackage[font=scriptsize]{caption}

\usepackage{hyperref}
\hypersetup{
    colorlinks=true,
    linkcolor=blue,
    filecolor=magenta,      
    urlcolor=cyan,
}

\newcommand{\R}{\mathbb{R}}
\newcommand{\X}{\mathcal{X}}

\newcommand{\E}{\mathbb{E}}

\newcommand{\calX}{\mathcal{X}}
\newcommand{\calC}{\mathcal{C}}
\newcommand{\calA}{\mathcal{A}}

\def\E{{\mathbb E}}

\def\R{{\mathbb R}}
\def\X{{\mathcal X}}

\def\pr{{\mbox{\rm Pr}}}

\newtheorem{thm}{Theorem}
\newtheorem{lemma}[thm]{Lemma}

\theoremstyle{definition}
\newtheorem{definition}{Definition}[section]
\newcommand{\mypara}[1]{\smallskip\noindent{\textbf{#1}}}

\begin{document}

%

%

\twocolumn[

\aistatstitle{A Non-Parametric Test to Detect Data-Copying in Generative Models}

\aistatsauthor{ Casey Meehan, Kamalika Chaudhuri, Sanjoy Dasgupta }

\aistatsaddress{ University of California, San Diego } ]

\begin{abstract}
Detecting overfitting in generative models is an important challenge in machine learning. In this work, we formalize a form of overfitting that we call {\em{data-copying}} -- where the generative model memorizes and outputs training samples or small variations thereof. We provide a three sample non-parametric test for detecting data-copying that uses the training set, a separate sample from the target distribution, and a generated sample from the model, and study the performance of our test on several canonical models and datasets.
\end{abstract}

\section{Introduction}
\begin{figure*}
    \centering
    \begin{subfigure}{.30\linewidth}
        \centering 
        \captionsetup{justification=centering}
        \includegraphics[width = 1.8in, angle = 90]{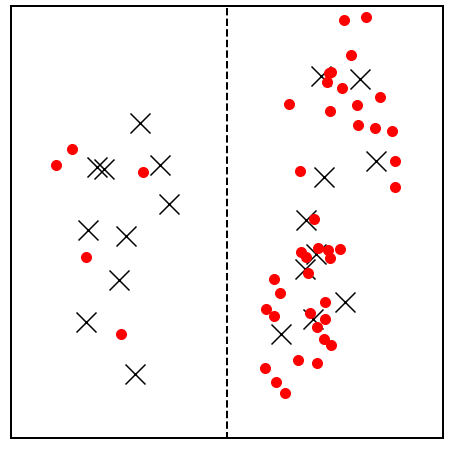}
        \caption{Illustration of over-/under-representation \\ Training sample: $\times$, Generated sample: \textcolor{red}{\textbullet}} \label{fig:cartoon overrep}
    \end{subfigure}
    \hspace{0.1in}
    \begin{subfigure}{.30\linewidth}
        \centering 
        \captionsetup{justification=centering}
        \includegraphics[width = 1.8in, angle = 90]{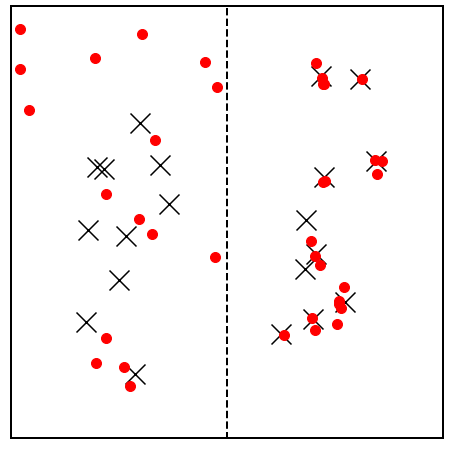}
        \caption{Illustration of data-copying/underfitting \\ Training sample: $\times$, Generated sample: \textcolor{red}{\textbullet}} \label{fig:cartoon data-copying}
    \end{subfigure}
        \hspace{0.1in}
    \begin{subfigure}{.32\linewidth}
        \centering
        \captionsetup{justification=centering}
        \includegraphics[height = 0.8in]{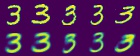}
        \vspace{0.1in}
        \includegraphics[height = 0.8in]{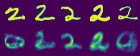}
        \caption{VAE copying/underfitting on MNIST \\ top: $Z_U = -8.54$, bottom: $Z_U = +3.30$} \label{fig:VAE mnist neighbors}
    \end{subfigure}
    \caption{Comparison of data-copying with over/under representation. Each image depicts a single instance space partitioned into two regions. Illustration \textbf{(a)} depicts an over-represented region (top) and under-represented region (bottom). This is the kind of overfitting evaluated by methods like FID score and Precision and Recall. Illustration \textbf{(b)} depicts a data-copied region (top) and underfit region (bottom). This is the type of overfitting focused on in this work. Figure \textbf{(c)} shows VAE-generated and training samples from a data-copied (top) and underfit (bottom) region of the MNIST instance space. In each 10-image strip, the bottom row provides random generated samples from the region and the top row shows their training nearest neighbors. Samples in the bottom region are on average further to their training nearest neighbor than held-out test samples in the region, and samples in the top region are closer, and thus `copying' (computed in embedded space, see Experiments section). }
    \label{fig:neighbor example}
\end{figure*} 
Overfitting is a basic stumbling block of any learning process. While it has been studied in great detail in the context of supervised learning, it has received much less attention in the unsupervised setting, despite being just as much of a problem.

To start with a simple example, consider a classical kernel density estimator (KDE), which given data $x_1, \ldots, x_n \in \R^d$, constructs a distribution over $\R^d$ by placing a Gaussian of width $\sigma > 0$ at each of these points, yielding the density
\begin{equation}\label{eq:kde}
q_{\sigma}(x) = \frac{1}{(2\pi)^{d/2}\sigma^d n} \sum_{i=1}^n  \exp\left( -\frac{\|x-x_i\|^2}{2 \sigma^2}\right) .
\end{equation}

The only parameter is the scalar $\sigma$. Setting it too small makes $q(x)$ too concentrated around the given points: a clear case of overfitting (see Appendix \textbf{Figure \ref{fig:half moon}}). This cannot be avoided by choosing the $\sigma$ that maximizes the log likelihood on the training data, since in the limit $\sigma \rightarrow 0$, this likelihood goes to $\infty$. 

The classical solution is to find a parameter $\sigma$ that has a low {\em{generalization gap}} -- that is, a low gap between the training log-likelihood and the log-likelihood on a held-out validation set. This method however often does not apply to the more complex generative models that have emerged over the past decade or so, such as Variational Auto Encoders (VAEs) \citep{kingma} and Generative Adversarial Networks (GANs)~\citep{goodfellow}. These models easily involve millions of parameters, and hence overfitting is a serious concern. Yet, a major challenge in evaluating overfitting is that these models do not offer exact, tractable likelihoods. VAEs can tractably provide a log-likelihood lower bound, while GANs have no accompanying density estimate at all. Thus any method that can assess these generative models must be based only on the samples produced. 


A body of prior work has provided tests for evaluating generative models based on samples drawn from them \citep{salimans, mehdi,  Ruslan_et_al, heusel}; however, the vast majority of these tests  focus on `mode dropping' and `mode collapse': the tendency for a generative model to either merge or delete high-density modes of the true distribution. A generative model that simply reproduces the training set or minor variations thereof will pass most of these tests. 

In contrast, this work formalizes and investigates a type of overfitting that we call `data-copying': the propensity of a generative model to recreate minute variations of a subset of training examples it has seen, rather than represent the true diversity of the data distribution. An example is shown in \textbf{Figure \ref{fig:cartoon data-copying}}; in the top region of the instance space, the generative model data-copies, or creates samples that are very close to the training samples; meanwhile, in the bottom region, it underfits. To detect this, we introduce a test that relies on three independent samples: the original training sample used to produce the generative model; a separate (held-out) test sample from the underlying distribution; and a synthetic sample drawn from the generator. 

Our key insight is that an overfit generative model would produce samples that are too close to the training samples -- closer on average than an independently drawn test sample from the same distribution. Thus, if a suitable distance function is available, then we can test for data-copying by testing whether the distances to the closest point in the training sample are on average smaller for the generated sample than for the test sample.

A further complication is that modern generative models tend to behave differently in different regions of space; a configuration as in \textbf{Figure \ref{fig:cartoon data-copying}} for example could cause a global test to fail. To address this, we use ideas from the design of non-parametric methods. We divide the instance space into cells, conduct our test separately in each cell, and then combine the results to get a sense of the average degree of data-copying.
 
Finally, we explore our test experimentally on a variety of illustrative data sets and generative models. Our results demonstrate that given enough samples, our test can successfully detect data-copying in a broad range of settings. 
 
\subsection{Related work}
\label{sec:better-global-test}

There has been a large body of prior work on the evaluation of generative models \citep{salimans,lopez, richardson, mehdi, Kilian, Ruslan_et_al} . Most are geared to detect some form of mode-collapse or mode-dropping: the tendency to either merge or delete high-density regions of the training data. Consequently, they fail to detect even the simplest case of extreme data-copying -- where a generative model memorizes and exactly reproduces a bootstrap sample from the training set. We discuss below a few such canonical tests.

To-date there is a wealth of techniques for evaluating whether a model mode-drops or -collapses. Tests like the popular Inception Score (IS), Frech\'et Inception Distance (FID) \citep{heusel}, Precision and Recall test \citep{mehdi}, and extensions thereof \citep{Kynk_improved, che_2016} all work by embedding samples using the features of a discriminative network such as `InceptionV3' and checking whether the training and generated samples are similar \emph{in aggregate}. The hypothesis-testing binning method proposed by \cite{richardson} also compares aggregate training and generated samples, but without the embedding step. The parametric Kernel MMD method proposed by \cite{gretton} uses a carefully selected kernel to estimate the distribution of both the generated and training samples and reports the maximum mean discrepancy between the two. All these tests, however, reward a generative model that only produces slight variations of the training set, and do not successfully detect even the most egregious forms of data-copying.

A test that can detect some forms of data-copying is the {\em{Two-Sample Nearest Neighbor}}, a non-parametric test proposed by~\cite{lopez}. Their method groups a training and generated sample of equal cardinality together, with training points labeled `1' and generated points labeled `0', and then reports the Leave-One-Out (LOO) Nearest-Neighbor (NN) accuracy of predicting `1's and `0's. Two values are then reported as discussed by \cite{Kilian} -- the leave-one-out accuracy of the training points, and the leave-one-out accuracy of the generated points. An ideal generative model should produce an accuracy of $0.5$ for each. More often, a mode-collapsing generative model will leave the training accuracy low and generated accuracy high, while a generative model that exactly reproduces the entire training set should produce zero accuracy for both. Unlike this method, our test not only detects exact data-copying, which is unlikely, but estimates whether a given model generates samples closer to the training set than it should, as determined by a held-out test set.

The concept of data-copying has also been explored by \cite{Kilian} (where it is called `memorization') for a variety of generative models and several of the above two-sample evaluation tests. Their results indicate that out of a variety of popular tests, only the two-sample nearest neighbor test is able to capture instances of extreme data-copying. 

\cite{gretton_2} explores three-sample testing, but for comparing the performance of different models, not for detecting overfitting. \cite{reviewer_paper} uses the three-sample test proposed by~\cite{gretton_2} for detecting data-copying; unlike ours, their test is global in nature. 


 Finally, other works concurrent with ours have explored parametric approaches to rooting out data-copying. A recent work by \cite{GAN_benchmarks} suggests that, given a large enough sample from the model, Neural Network Divergences are sensitive to data-copying. In a slightly different vein, a recent work by \cite{latent_recovery} investigates whether latent-parameter models memorize training data by learning the reverse mapping from image to latent code. The present work departs from those by offering a probabilistically motivated non-parametric test that is entirely model agnostic.



\section{Preliminaries}

We begin by introducing some notation and formalizing the definitions of overfitting. Let $\X$ denote an instance space in which data points lie, and $P$ an unknown underlying distribution on this space. A training set $T$ is drawn from $P$ and is used to build a generative model $Q$. We then wish to assess whether $Q$ is the result of overfitting: that is, whether $Q$ produces samples that are too close to the training data. To help ascertain this, we are able to draw two additional samples:
\begin{itemize}
\item A fresh sample of $n$ points from $P$; call this $P_n$.
\item A sample of $m$ points from $Q$; call this $Q_m$.
\end{itemize}
As illustrated in \textbf{Figures \ref{fig:cartoon overrep}, \ref{fig:cartoon data-copying}}, a generative model can overfit locally in a region $\calC \subseteq \calX$. To characterize this, for any distribution $D$ on $\calX$, we use $D|_{\calC}$ denote its restriction to the region $\calC$, that is, 
\begin{align*}
    D|_{\calC}(\calA) = \frac{D(\calA \cap \calC)}{D(\calC)} 
    \quad \text{for any $\calA \subseteq \X$.}
\end{align*}

\subsection{Definitions of Overfitting}
\label{sec:types of overfitting}
We now formalize the notion of data-copying, and illustrate its distinction from other types of overfitting. 

Intuitively, data-copying refers to situations where $Q$ is ``too close'' to the training set $T$; that is, closer to $T$ than the target distribution $P$ happens to be. We make this quantitative by choosing a distance function $d: \X \rightarrow \R$ from points in $\X$ to the training set, for instance, $d(x) = \min_{t \in T} \|x - t\|^2$, if $\X$ is a subset of Euclidean space. 

Ideally, we desire that $Q$'s expected distance to the training set is the same as that of $P$'s, namely $\E_{X \sim P}[d(X)] = \E_{Y \sim Q}[d(Y)]$. We may rewrite this as follows: given any distribution $D$ over $\calX$, define $L(D)$ to be the one-dimensional distribution of $d(X)$ for $X \sim D$. We consider data-copying to have occurred if random draws from $L(P)$ are systematically larger than from $L(Q)$. The above equalized expected distance condition can be rewritten as
\begin{align}
      \E_{Y \sim Q}[d(Y)] - \E_{X \sim P}[d(X)] 
      &= \E_{^{A \sim L(P)}_{B \sim L(Q)}}[B - A] 
      = 0
      \label{eqn:equal expected distance}
\end{align}
However, we are less interested in how \emph{large} the difference is, and more in how \emph{often} $B$ is larger than $A$. Let 
\begin{align*}
    \Delta_T(P,Q) &= \pr \left( B > A \ \big|\  B \sim L(Q), A \sim L(P) \right)
\end{align*}
where $0 \leq \Delta_T(P,Q) \leq 1$ represents how `far' $Q$ is from training sample $T$ as compared to true distribution $P$. A more interpretable yet equally meaningful condition is
\begin{align*}
    \Delta_T(P,Q) 
    &= \E_{^{A \sim L(P)}_{B \sim L(Q)}}[\mathds{1}_{B > A}] 
    \approx 
    \frac{1}{2} 
\end{align*}
which guarantees \eqref{eqn:equal expected distance} if densities $L(P)$ and $L(Q)$ have the same shape, but could plausibly be mean-shifted. 

If $\Delta_T(P,Q) \ll \frac{1}{2}$, $Q$ is data-copying training set $T$, since samples from $Q$ are systematically closer to $T$ than are samples from $P$. However, even if $\Delta_T(P,Q) \geq \frac{1}{2}$, $Q$ may still be data-copying. As exhibited in \textbf{Figures \ref{fig:cartoon data-copying}} and \textbf{\ref{fig:VAE mnist neighbors}}, a model $Q$ may data-copy in one region and underfit in others. In this case, $Q$ may be further from $T$ than is $P$ \emph{globally}, but much closer to $T$ \emph{locally}. As such, we consider $Q$ to be data-copying if it is overfit in a subset $\calC \subseteq \calX$:

\begin{definition}[Data-Copying]
    A generative model $Q$ is \textit{data-copying} training set $T$ if, in some region $\calC \subseteq \calX$, it is systematically closer to $T$ by distance metric $d:\calX \rightarrow \R$ than are samples from $P$ . Specifically, if
    \begin{align*}
        \Delta_T(P|_{\calC}, Q|_{\calC}) < \frac{1}{2}
    \end{align*}
\end{definition}

Observe that data-copying is orthogonal to the type of overfitting addressed by many previous works \citep{heusel, mehdi}, which we call `over-representation'. There, $Q$ overemphasizes some region of the instance space $\calC \subseteq \calX$, often a region of high density in the training set $T$. For the sake of completeness, we provide a formal definition below.


\begin{definition}[Over-Representation]
    A generative model $Q$ is \textit{over-representing} $P$ in some region $\calC \subseteq \calX$, if the probability of drawing $Y \sim Q$ is much greater than it is of drawing $X \sim P$. Specifically, if 
    \begin{align*}
        Q(\calC) - P(\calC) \gg 0 
    \end{align*}
\end{definition}

Observe that it is possible to over-represent without data-copying and vice versa. For example, if $P$ is an equally weighted mixture of two Gaussians, and $Q$ perfectly models one of them, then $Q$ is over-representing without data-copying. On the other hand, if $Q$ outputs a bootstrap sample of the training set $T$, then it is data-copying without over-representing. The focus of the rest of this work is on data-copying.  

\section{A Test For Data-Copying}

Having provided a formal definition, we next propose a hypothesis test to detect data-copying.

\subsection{A Global Test}
\label{sec:simple-global-test}
We introduce our data-copying test in the global setting, when $\calC = \calX$. Our null hypothesis $H_0$ suggests that $Q$ may equal $P$: 
\begin{align}
    \label{eqn: null hypothesis}
     H_0: \Delta_T(P,Q) \ = \ \frac{1}{2}
\end{align}
There are well-established non-parametric tests for this hypothesis, such as the Mann-Whitney $U$ test \citep{mannwhitney}. Let $A_i \sim L(P_n), B_j \sim L(Q_m)$ be samples of $L(P), L(Q)$ given by $P_n, Q_m$ and their distances $d(X)$ to training set $T$. The $U$ statistic estimates the probability in Equation \ref{eqn: null hypothesis} by measuring the number of all $mn$ pairwise comparisons in which $B_j > A_i$. An efficient and simple method to gather and interpret this test is as follows: 
\begin{enumerate}
    \item Sort the $n+m$ values $L(P_n) \cup L(Q_m)$ such that each instance $l_i \in L(P_n), l_j \in L(Q_m)$ has rank $R(l_i), R(l_j)$, starting from rank 1, and ending with rank $n+m$. $L(P_n), L(Q_m)$ have no tied ranks with probability 1 assuming their distributions are continuous.
    \item Calculate the rank-sum for $L(Q_m)$ denoted $R_{Q_m}$, and its $U$ score denoted $U_{Q_m}$:  
    \begin{align*}
       R_{Q_m} = \sum_{l_j \in L(Q_m)} R(l_j), \quad 
       U_{Q_m} = R_{Q_m} - \frac{m(m+1)}{2}
    \end{align*}
    Consequently, $U_{Q_m} = \sum_{ij} \mathds{1}_{B_j > A_i}$. 
    \item Under $H_0$, $U_{Q_m}$ is approximately normally distributed with $>20$ samples in both $L(Q_m)$ and $L(P_n)$, allowing for the following $z$-scored statistic
    \begin{align*}
        Z_U\big(L(P_n), L(Q_m);T\big) = \frac{U_{Q_m} - \mu_U}{\sigma_U}, \\
        \mu_U = \frac{mn}{2}, \quad
        \sigma_U = \sqrt{\frac{mn(m + n + 1)}{12}}
    \end{align*}
\end{enumerate}
$Z_U$ provides us a data-copying statistic with normalized expectation and variance under $H_0$. $Z_U \ll 0$ implies data-copying, $Z_U \gg 0$ implies underfitting. $Z_U < -5$ implies that if $H_0$ holds, $Z_U$ is as likely as sampling a value $< -5$ from a standard normal.

Observe that this test is completely model agnostic and uses no estimate of likelihood. It only requires a meaningful distance metric, which is becoming common practice in the evaluation of mode-collapse and -dropping \citep{heusel, mehdi} as well.

\subsection{Handling Heterogeneity}
\label{sec:local-versus-global}
As described in Section \ref{sec:types of overfitting}, the above global test can be fooled by generators $Q$ which are very close to the training data in some regions of the instance space (overfitting) but very far from the training data in others (poor modeling). 

We handle this by introducing a {\it local} version of our test. Let $\Pi$ denote any partition of the instance space $\calX$, which can be constructed in any manner. In our experiments, for instance, we run the $k$-means algorithm on $T$, so that $|\Pi| = k$. As the number of training and test samples grows, we may increase $k$ and thus the instance-space resolution of our test. Letting $L_\pi(D) = L(D|_{\pi})$ be the distribution of distances-to-training-set within cell $\pi \in \Pi$, we probe each cell of the partition $\Pi$ individually. 

\paragraph{Data Copying.}
\label{sec:data copying statistic}
To offer a summary statistic for data copying, we collect the $z$-scored Mann-Whitney $U$ statistic, $Z_U$, described in Section \ref{sec:simple-global-test} in each cell $\pi$. Let $P_n(\pi) = |\{x: x \in P_n, x \in \pi\}| / n$ denote the fraction of $P_n$ points lying in cell $\pi$, and similarly for $Q_m(\pi)$. The $Z_U$ test for cell $\pi$ and training set $T$ will then be denoted as $Z_U\big(L_\pi(P_n), L_\pi(Q_m); T\big)$, where $L_\pi(P_n) = \{d(x): x \in P_n, x \in \pi\}$ and similarly for $L_\pi(Q_m)$. See \textbf{Figure \ref{fig:VAE mnist neighbors}} for examples of these in-cell scores. For stability, we only measure data-copying for those cells significantly represented by $Q$, as determined by a threshold $\tau$. Let $\Pi_{\tau}$ be the set of all cells in the partition $\Pi$ for which $Q_m(\pi) \geq \tau$. Then, our summary statistic for data copying averages across all cells represented by $Q$:  

\begin{align*}
    C_T(P_n, Q_m) \coloneqq  \frac{\sum_{\pi \in \Pi_\tau} P_n(\pi) Z_U\big(L_\pi(P_n), L_\pi(Q_m); T\big)}{ \sum_{\pi \in \Pi_\tau} P_n(\pi) }
\end{align*}

\paragraph{Over-Representation.}
\label{sec:over rep statistic}
The above test will not catch a model that heavily over- or under-represents cells. For completeness, we next provide a simple representation test that is essentially used by \cite{richardson}, now with an independent test set instead of the training set.  

With $n, m\geq 20$ in cell $\pi$, we may treat $Q_m(\pi), P_n(\pi)$ as Gaussian random variables. We then check the null hypothesis $H_0 : 0 = P(\pi) - Q(\pi)$. Assuming this null hypothesis, a simple $z$-test is: 
\begin{align*}
Z_\pi = \frac{Q_m(\pi) - P_n(\pi)}
 {\sqrt{ \widehat{p}\big(1 - \widehat{p}\big) \big( \frac{1}{n} + \frac{1}{m} \big) }}
\end{align*}
where $\widehat{p} = \frac{nP_n(\pi) + mQ_m(\pi)}{n + m}$. We then report two values for a significance level $s = 0.05$: the number of significantly different cells (`bins') with $Z_\pi > s$ (NDB over-representing), and the number with $Z_\pi < -s$ (NDB under-representing).

Together, these summary statistics --- $C_T$, NDB-over, NDB-under --- detect the ways in which $Q$ broadly represents $P$ without directly copying the training set $T$. 

\subsection{Performance Guarantees}
\label{sec:perf guarantees}

We next provide some simple guarantees on the performance of the global test statistic $U(Q_m)$. Guarantees for the average test is more complicated, and is left as a direction for future work.

We begin by showing that when the null hypothesis $H_0$ does not hold, $U_{Q_m}$ has some desirable properties -- $\frac{1}{mn} U_{Q_m}$ is a consistent estimator of the quantity of interest, $\Delta_T(P,Q)$: 
\begin{thm}
    \label{thm:consistent}
    For true distribution $P$, model distribution $Q$, and distance metric $d:\calX \rightarrow \R$, the estimator $\frac{1}{mn} U_{Q_m} \rightarrow_P \Delta(P,Q)$ according to the concentration inequality 
    \begin{align*}
        \pr \big( \ \big|\frac{1}{mn} U_{Q_m} - \Delta(P,Q) \big| \geq t\big) \leq 
        \exp \bigg( -\frac{2 t^2 mn}{m + n} \bigg)
    \end{align*}
\end{thm}
Furthermore, when the model distribution $Q$ actually matches the true distribution $P$, under modest assumptions we can expect $\frac{1}{mn} U_{Q_m}$ to be near $\frac{1}{2}$: 
\begin{thm}
    \label{thm:fallback}
    If $Q = P$, and the corresponding distance distribution $L(Q) = L(P)$ is non-atomic, then
    \begin{align*}
         \E \Big[ \frac{1}{mn}U_{Q_m} \Big] &= \frac{1}{2}
         \quad \text{and} \quad \E [Z_U] = 0
    \end{align*}
\end{thm} 
Proofs are provided in Appendices \ref{sec:proof consistent} and \ref{sec:proof fallback}. 

Additionally, we show that for a Gaussian Kernel Density Estimator, the parameter $\sigma$ that satisfies the condition in Equation \ref{eqn:equal expected distance} is the $\sigma$ corresponding to a maximum likelihood Gaussian KDE model. Recall that a KDE model is described by
\begin{equation}
q_\sigma(x) = \frac{1}{(2\pi)^{k/2}|T| \sigma^k} \sum_{t \in T} \exp\left( -\frac{\|x-t\|^2}{2 \sigma^2}\right) ,
\label{eqn:kde}
\end{equation}
where the posterior probability that a random draw $x \sim q_\sigma(x)$ comes from the Gaussian component centered at training point $t$ is
\begin{align*}
    Q_\sigma(t|x) = \frac{\exp(-\|x - t\|^2/(2 \sigma^2))}{\sum_{t' \in T}\exp(-\|x - t'\|^2/(2 \sigma^2))}
\end{align*}

\begin{lemma}
For the kernel density estimator (\ref{eqn:kde}), the maximum-likehood choice of $\sigma$, namely the maximizer of $\E_{X \sim P}[\log q_\sigma(X)]$, satisfies
\begin{align*}
	\E_{X \sim P} \bigg[ \sum_{t \in T} Q_\sigma(t|X) &\|X - t\|^2 \bigg] = \\
	&\E_{Y \sim Q_\sigma} \bigg[ \sum_{t \in T} Q_\sigma(t|Y) \|Y - t\|^2 \bigg]
\end{align*}
\label{lemma:kde}
\end{lemma}
See Appendix \ref{sec:appendix kde lemma} for proof.
Unless $\sigma$ is large, we know that for any given $x \in \calX$, $\sum_{t \in T} Q_\sigma(t|x) \|x - t\|^2 \approx d(x) = \min_{t \in T} \|x - t\|^2 $. So, enforcing that $\E_{X \sim P}[d(X)] = \E_{Y \sim Q}[d(Y)]$, and more loosely that $\E_{^{A \sim L(P)}_{B \sim L(Q)}}[\mathds{1}_{B > A}] = \frac{1}{2}$ provides an excellent non-parametric approach to selecting a Gaussian KDE, and ought to be enforced for any $Q$ attempting to emulate $P$; after all, Theorem \ref{thm:fallback} points out that effectively any model with $Q = P$ also yields this condition.

\section{Experiments}
\label{sec:experiments}
\begin{figure*}
    \centering
    \begin{subfigure}{.24\linewidth}
        \centering
        \includegraphics[width = 1\linewidth]{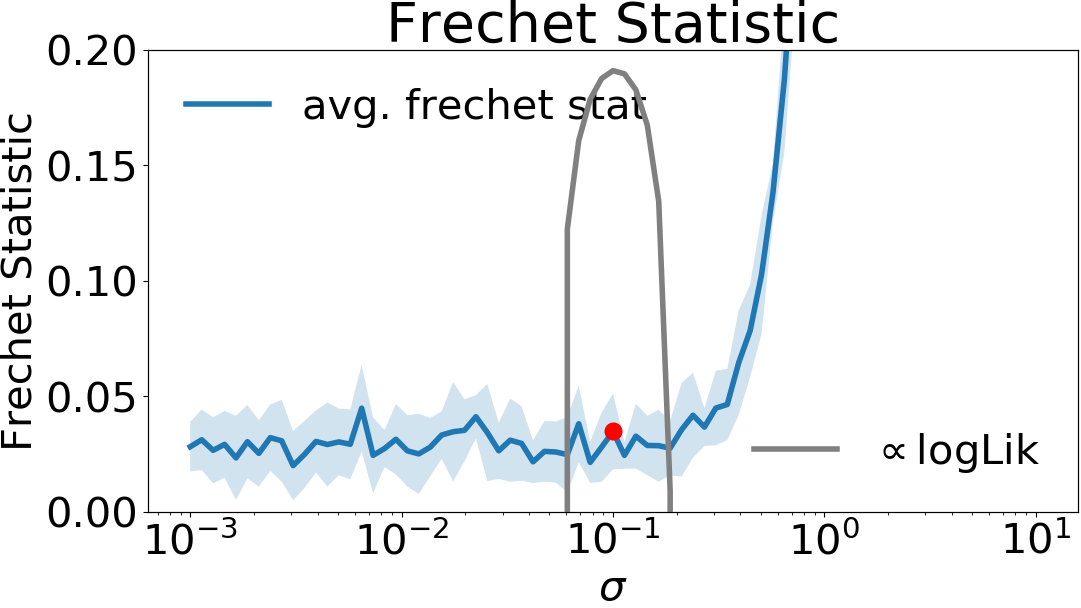}
        \caption{}\label{fig:moons frechet}
    \end{subfigure}
    \begin{subfigure}{0.24\linewidth}
        \centering
        \includegraphics[width = 1\linewidth]{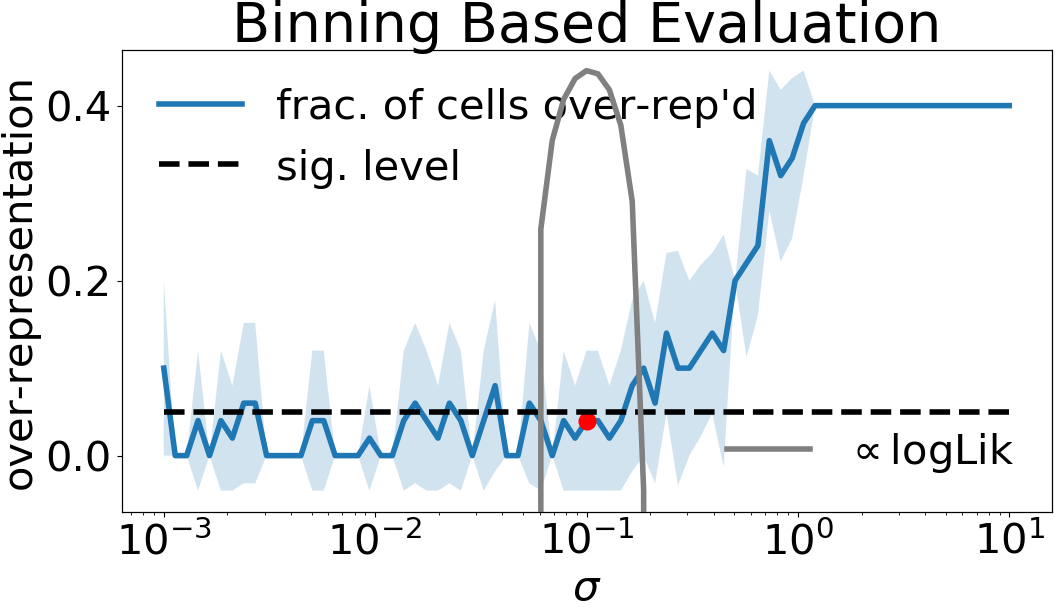}
        \caption{}\label{fig:moons binning}
    \end{subfigure}
    \begin{subfigure}{.24\linewidth}
        \centering
        \includegraphics[width = 1\linewidth]{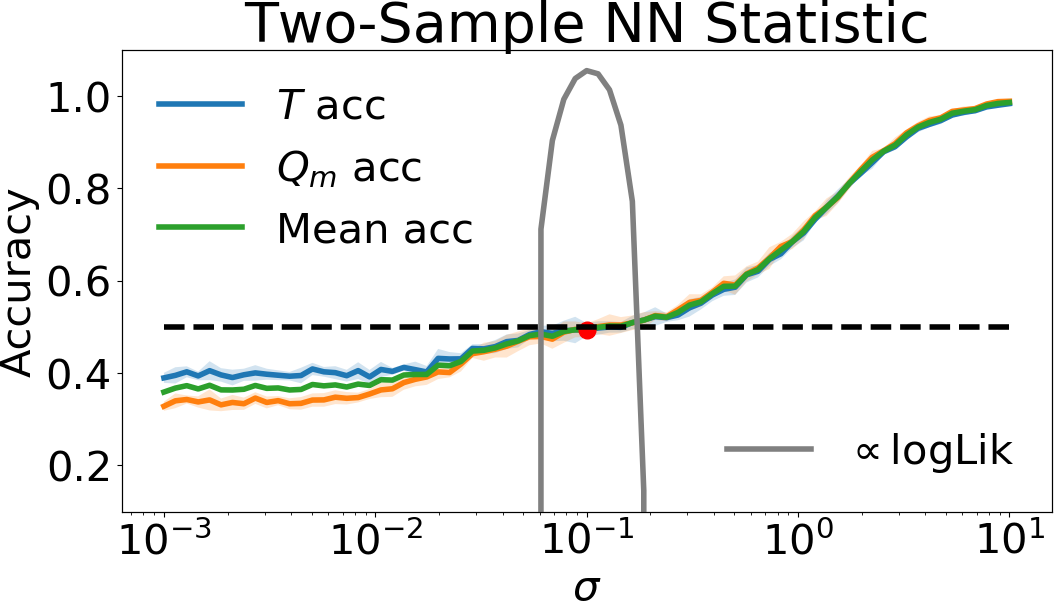}
        \caption{}\label{fig:moons NN compare}
    \end{subfigure}
    \begin{subfigure}{0.24\linewidth}
        \centering
        \includegraphics[width = 1\linewidth]{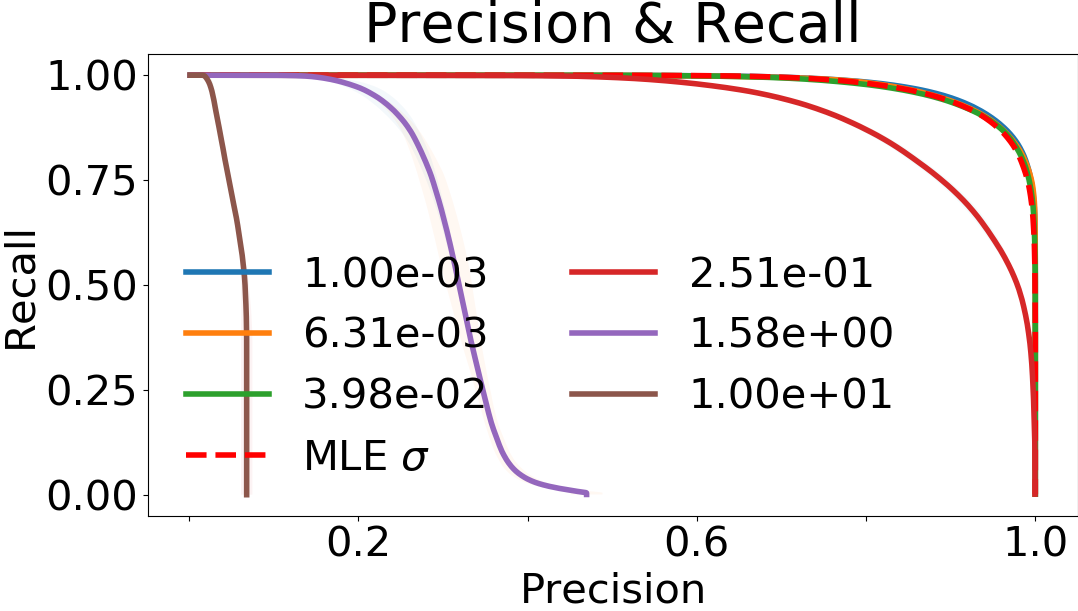}
        \caption{}\label{fig:moons PR}
    \end{subfigure}
    \caption{Response of four baseline test methods to data-copying of a Gaussian KDE on `moons' dataset. Only the two-sample NN test \textbf{(c)} is able to detect data-copying KDE models as $\sigma$ moves below $\sigma_{\text{MLE}}$ (depicted as a red dot). The gray trace is proportional to the KDE's log-likelihood measured on a held-out validation set.}
    \label{fig:compare methods}
\end{figure*} 

Having clarified what we mean by data-copying in theory, we turn our attention to data copying by generative models in practice\footnote{\href{https://github.com/casey-meehan/data-copying}{\texttt{https://github.com/casey-meehan/data-copying}}}. We leave representation test results for the appendix, since this behavior has been well studied in previous works. Specifically, we aim to answer the two following questions:
\begin{enumerate}
    \item  Are the existing tests that measure generative model overfitting able to capture data-copying? 
    \item As popular generative models range from over- to underfitting, does our test indicate data-copying, and if so, to what degree? 
\end{enumerate}

\paragraph{Training, Generated and Test Sets.}
In all of the following experiments, we select a training dataset $T$ with test split $P_n$, and a generative model $Q$ producing a sample $Q_m$. We perform $k$-means on $T$ to determine partition $\Pi$, with the objective of having a reasonable population of both $T$ and $P_n$ in each $\pi \in \Pi$. We set threshold $\tau$, such that we are guaranteed to have at least 20 samples in each cell in order to validate the gaussian assumption of $Z_\pi, Z_U$. 



\begin{figure*}
    \centering
    \begin{subfigure}{.27\linewidth}
        \centering
        \includegraphics[width = 1\linewidth]{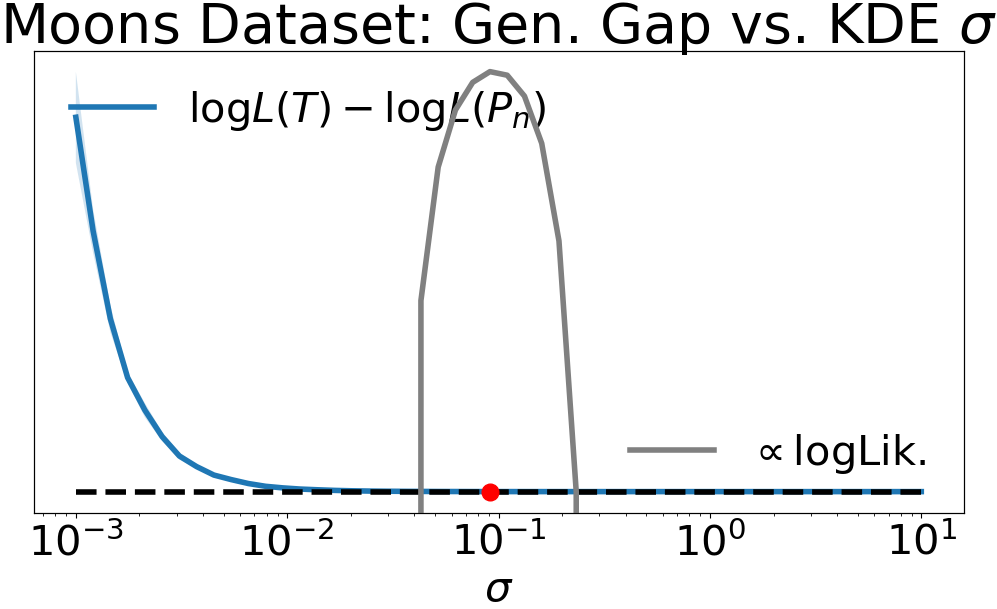}
        \caption{}\label{fig:moons gen gap}
    \end{subfigure}
        \hfill 
    \begin{subfigure}{.27\linewidth}
        \centering
        \includegraphics[width = 1\linewidth]{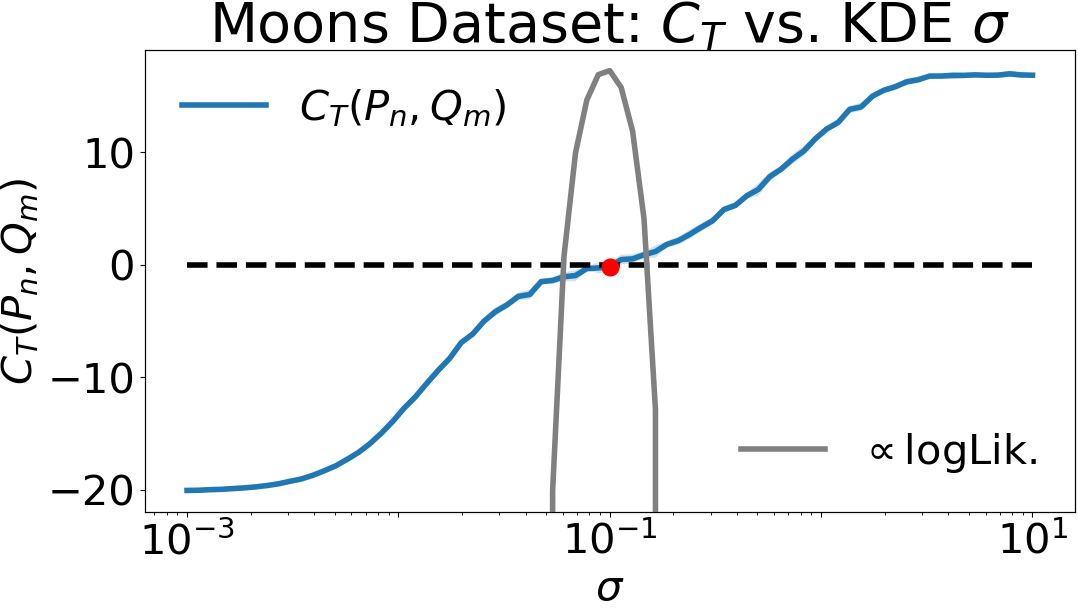}
        \caption{}\label{fig:moons z score}
    \end{subfigure}
        \hfill
    \begin{subfigure}{.27\linewidth}
        \centering
        \includegraphics[width = 1\linewidth]{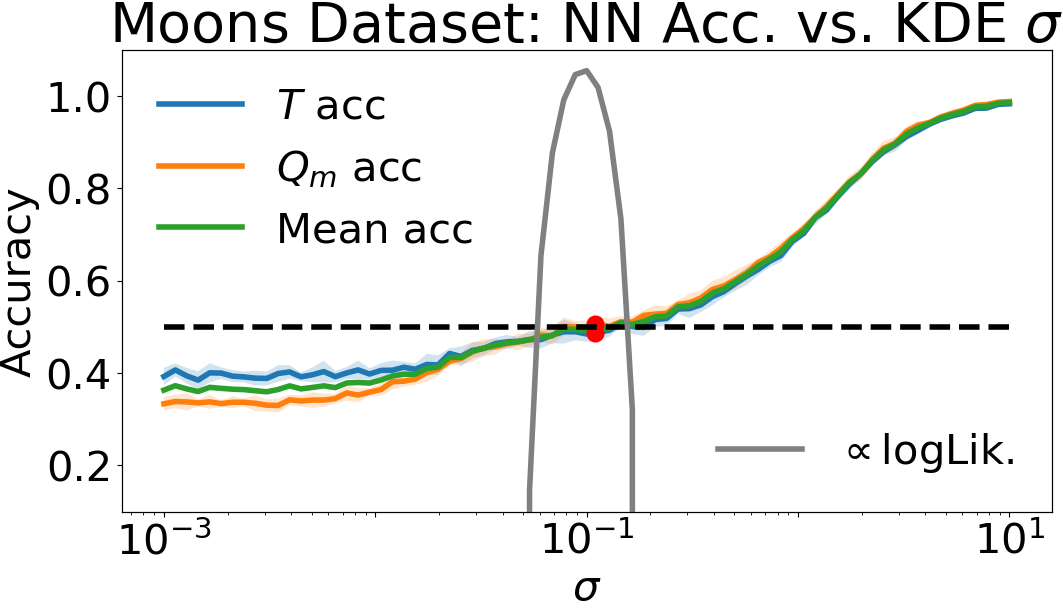}
        \caption{}\label{fig:moons NN}
    \end{subfigure} 
        \vskip\baselineskip
    \begin{subfigure}{.27\linewidth}
        \centering
        \includegraphics[width = 1\linewidth]{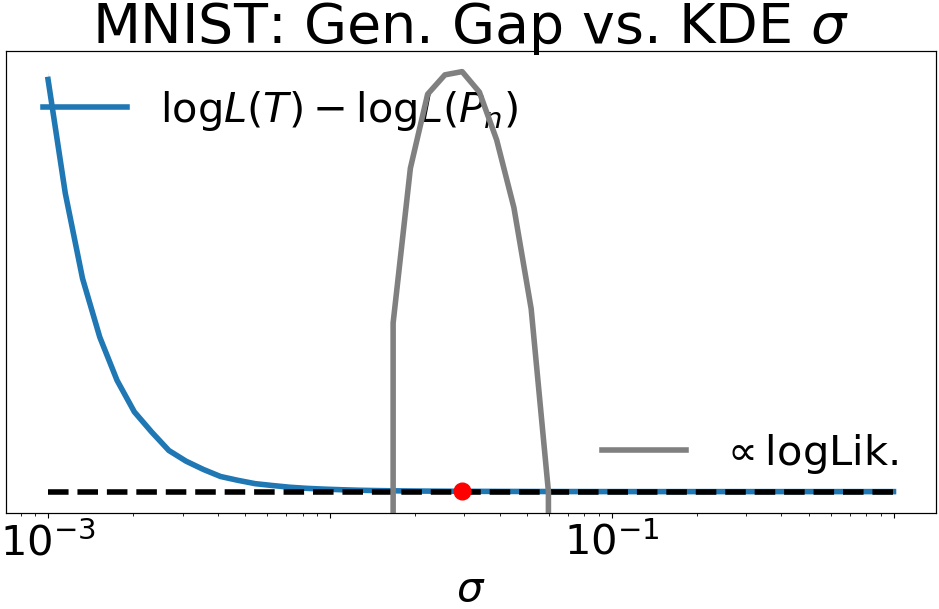}
        \caption{}\label{fig:mnist gen gap}
    \end{subfigure}
        \hfill 
    \begin{subfigure}{.27\linewidth}
        \centering
        \includegraphics[width = 1\linewidth]{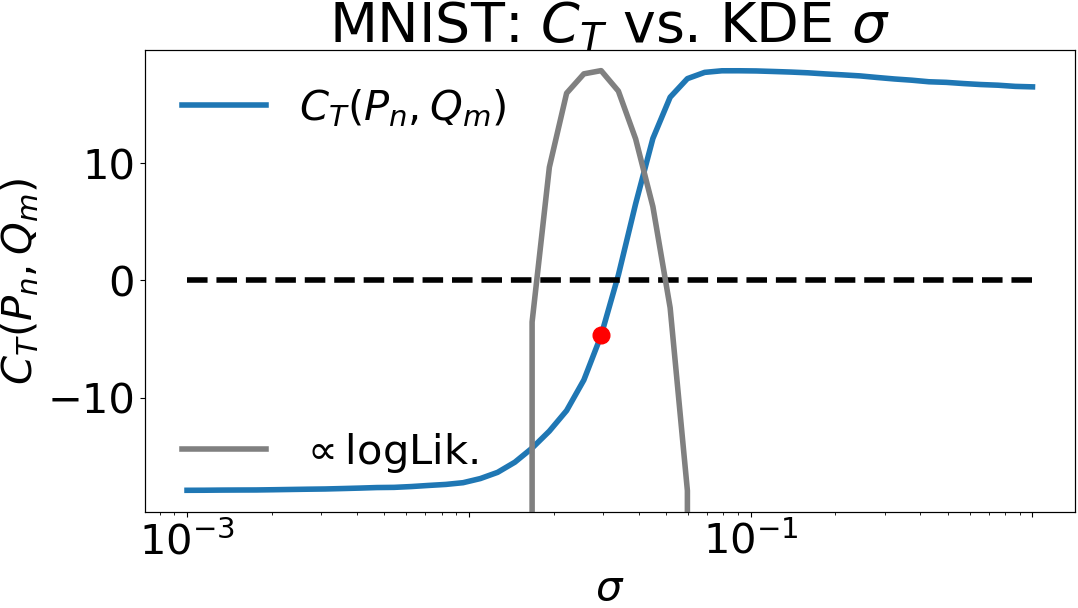}
        \caption{}\label{fig:mnist kde z score}
    \end{subfigure}
        \hfill
    \begin{subfigure}{.27\linewidth}
        \centering
        \includegraphics[width = 1\linewidth]{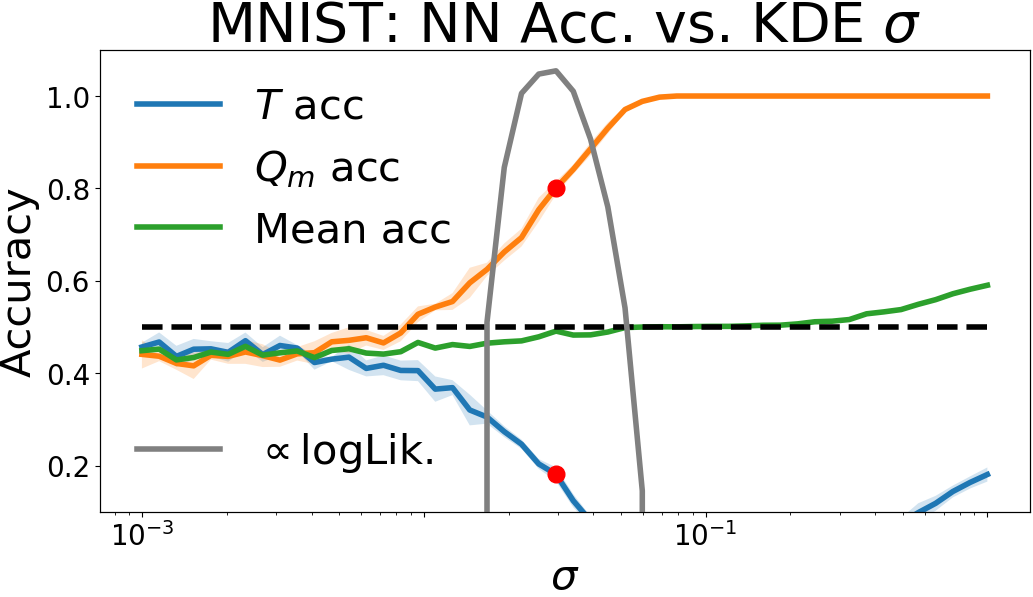}
        \caption{}\label{fig:mnist kde NN}
    \end{subfigure}
    \caption{$C_T(P_n, Q_m)$ vs. NN baseline and generalization gap on moons and MNIST digits datasets. \textbf{(a,b,c)} compare the three methods on the moons dataset. \textbf{(d,e,f)} compare the three methods on MNIST. In both data settings, the $C_T$ statistic is far more sensitive to the data-copying regime $\sigma \ll \sigma_{\text{MLE}}$ than the NN baseline. It is more sensitive to underfitting $\sigma \gg \sigma_{\text{MLE}}$ than the generalization gap test. The red dot denotes $\sigma_{\text{MLE}}$, and the gray trace is proportional to the KDE's log-likelihood measured on a held-out validation set.}
    \label{fig:KDE results}
\end{figure*}

\subsection{Detecting data-copying}
\label{sec:sensitivity to data-copying}

First, we investigate which of the existing generative model tests can detect explicit data-copying.

\paragraph{Dataset and Baselines.} For this experiment, we use the simple two-dimensional `moons' dataset, as it affords us limitless training and test samples and requires no feature embedding (see Appendix \ref{sec:appendix moons kDE} for an example). 

As baselines, we probe four of the methods described in our Related Work section to see how they react to data-copying: two-sample NN \citep{lopez}, FID \citep{heusel}, Binning-Based Evaluation \citep{richardson}, and Precision \& Recall \citep{mehdi}. A detailed description of the methods is provided in Appendix \ref{sec:appendix moons experiments}. Note that, without an embedding, FID is simply the Frech\'et distance between two maximum likelihood normal distributions fit to $T$ and $Q_m$. We use the same size generated and training sample for all methods. Note that the two-sample NN test requires the generated sample size $m$ to be equal to the training sample size $T$. When $m < |T|$ (especially for large datasets and computationally burdensome samplers) we use an $m$-size training subsample $\widetilde{T}$. 


\paragraph{Experimental Methodology.} We choose as our generative model $Q$ a Gaussian KDE as it allows us to force explicit data-copying by setting $\sigma$ very low. As $\sigma \rightarrow 0$, $Q$ becomes a bootstrap sampler of the original training set. If a given test method can detect the level of data-copying by $Q$ on $T$, it will provide a different response to a heavily over-fit KDE $Q$ ($\sigma \ll \sigma_{\text{MLE}}$), a well-fit KDE $Q$ ($\sigma \approx \sigma_{\text{MLE}}$), and an underfit KDE $Q$ ($\sigma \gg \sigma_{\text{MLE}}$). 

\paragraph{Results.} \textbf{Figure \ref{fig:compare methods}} depicts how each baseline method responds to KDE $Q$ models of varying degrees of data-copying, as $Q$ ranges from data-copying ($\sigma = 0.001$) up to heavily underfit ($\sigma = 10$). The Frech\'et and Binning methods report effectively the same value for all $\sigma \leq \sigma_{\text{MLE}}$, indicating inability to detect data-copying. Similarly, the Precision-Recall curves for different $\sigma$ values are nearly identical for all $\sigma \leq \sigma_{\text{MLE}}$, and only change for large $\sigma$. 

The two-sample NN test does show a mild change in response as $\sigma$ decreases below $\sigma_{\text{MLE}}$. This makes sense; as points in $Q_m$ become closer to points in $T$, the two-sample NN accuracy should steadily decline. The primary reason it does not drop to zero is due to the $m$ subsampled training points, $\widetilde{T} \subset T$, needed to perform this test. As such, each training point $t \in T$ being copied by generated point $q \in Q_m$ is unlikely to be present in $\widetilde{T}$ during the test. This phenomenon is especially pronounced in some of the following settings. 
	However, even when $m = |T|$, this test will not reduce to zero as $\sigma \rightarrow 0$ due to the well-known result that a bootstrap sample of $T$ will only include $\approx 1 - \nicefrac{1}{e}$ of the samples in $T$. Consequently, several training samples will not have a generated sample as nearest neighbor. The $C_T$ test avoids this by specifically finding the training nearest neighbor of each generated sample. 

The reason most of these tests fail to detect data-copying is because most existing methods focus on another type of overfitting: mode-collapse and -dropping, wherein entire modes of $P$ are either forgotten or averaged together. However, if a model begins to data-copy, it is definitively overfitting \emph{without} mode-collapsing.

Note that the above four baselines are all two sample tests that do not use $P_n$ as $C_T$ does. For completeness, we present experiments with an additional, three sample baseline in Appendix \ref{sec:kMMD}. Here, we repeat the `moons' dataset experiment with the three-sample kernel MMD test originally proposed by \cite{gretton_2} for generative model selection and later adapted by \cite{reviewer_paper} for testing model over-fitting. We observe in \textbf{Figure \ref{fig:kMMD comparison moons kMMD}} that the three-sample kMMD test does not detect data-copying, treating the MLE model similarly to overfit models with $\sigma << \sigma_{\text{MLE}}$. See Appendix \ref{sec:kMMD} for experimental details.

\begin{figure*}[h]
    \centering
    \begin{subfigure}{.27\linewidth}
        \centering
        \includegraphics[width = 1\linewidth]{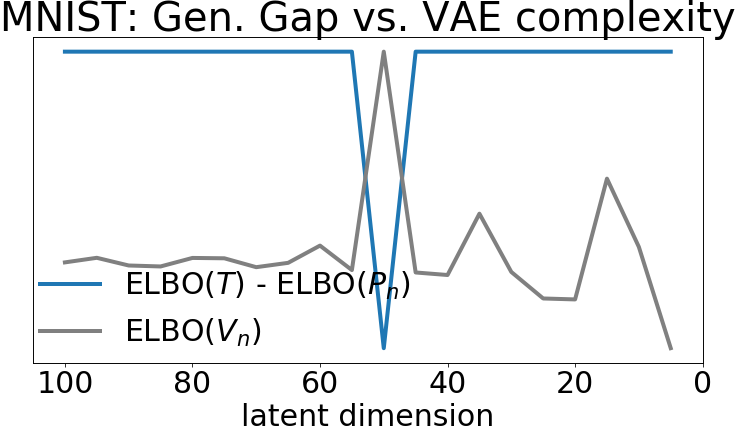}
        \caption{}\label{fig:vae gen gap}
    \end{subfigure}
    \hfill 
    \begin{subfigure}{.27\linewidth}
        \centering
        \includegraphics[width = 1\linewidth]{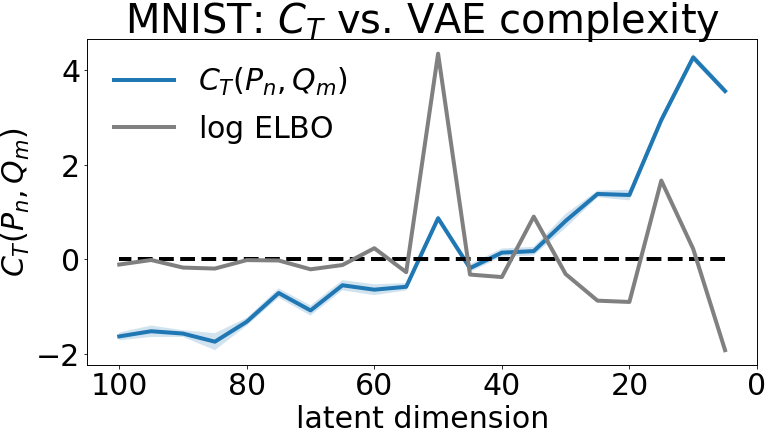}
        \caption{}\label{fig:vae C_T vs d}
    \end{subfigure}
        \hfill
    \begin{subfigure}{.27\linewidth}
        \centering
        \includegraphics[width = 1\linewidth]{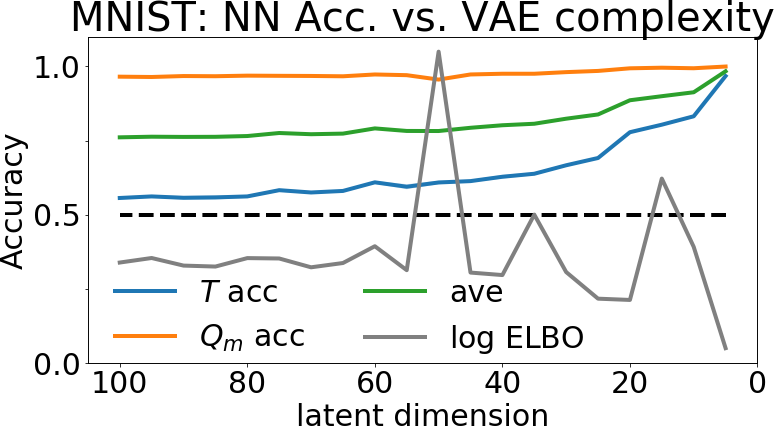}
        \caption{}\label{fig:vae NN vs d}
    \end{subfigure}
        \hfill
    \begin{subfigure}{.27\linewidth}
        \centering
        \includegraphics[width = 1\linewidth]{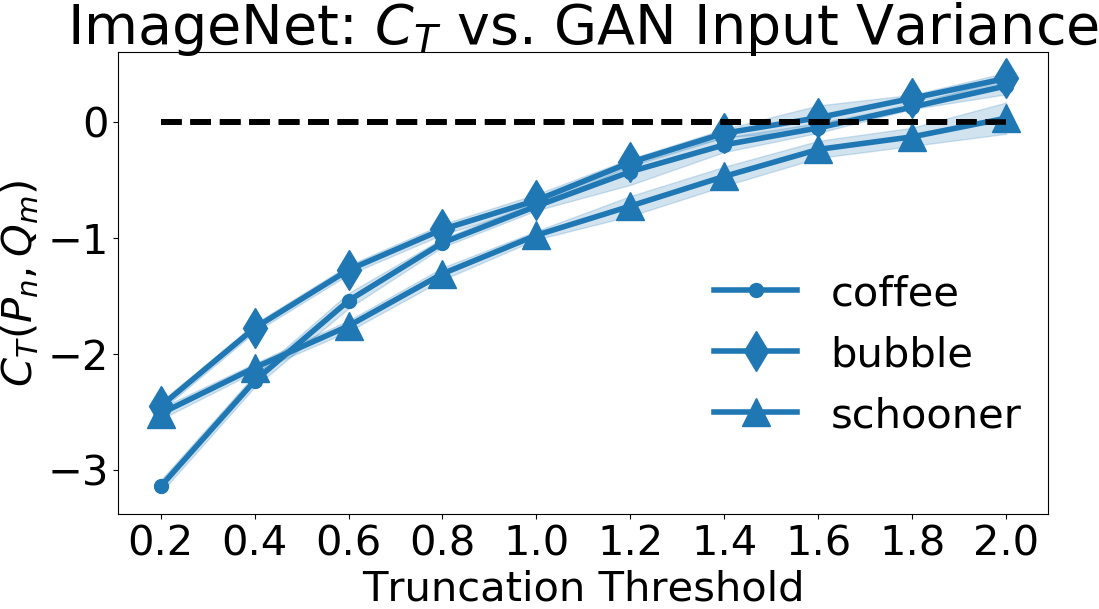}
    \caption{}\label{fig:biggan C_T score}
    \end{subfigure}
        \quad 
    \begin{subfigure}{.27\linewidth}
        \centering
        \includegraphics[width = 1\linewidth]{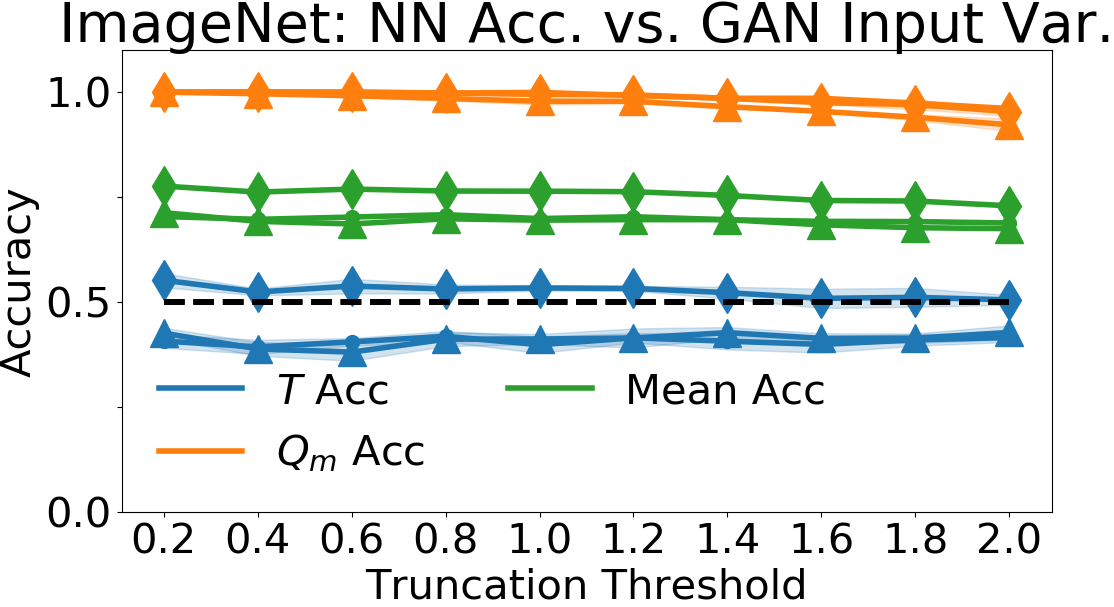}
        \caption{}\label{fig:biggan NN}
    \end{subfigure}
    \caption{Neural model data-copying: figures \textbf{(b)} and \textbf{(d)} demonstrate the $C_T$ statistic identifying data-copying in an MNIST VAE and ImageNet GAN as they range from heavily over-fit to underfit. \textbf{(c)} and \textbf{(e)} demonstrate the relative insensitivity of the NN baseline to this overfitting, as does figure \textbf{(a)} of the generalization (ELBO) gap method for VAEs. (Note, the markers for \textbf{(d)} apply to the traces of \textbf{(e)})}
    \label{fig:neural experiments}
\end{figure*}

\subsection{Measuring degree of data-copying}
\label{sec:degree of data-copying}
We now aim to answer the second question raised at the beginning of this section: does our test statistic $C_T(P_n, Q_m)$ detect and quantify data-copying? 

We focus on three generative models: Gaussian KDEs, Variational Autoencoders (VAEs) and Generative Adversarial Networks (GANs). For these experiments, we consider two baselines in addition to our method --- the two-sample NN test, and the likelihood generalization gap where it can be computed or approximated. 

\subsubsection{KDE-based tests}
\label{sec:KDE experiments}
 
First we consider Gaussian KDEs. While KDEs do not provide a reliable likelihood in high dimension \citep{theis}, they do have several advantages as a preliminary benchmark -- they allow us to directly force data-copying, and can help investigate the practical implications of the theoretical connection between the maximum likelihood KDE and $C_T \approx 0$ as described in Lemma \ref{lemma:kde}. We explore two datasets with Gaussian KDE: the 'moons' dataset, and MNIST. 

\paragraph{KDEs: `moons' dataset.} Here, we repeat the experiment performed in Section \ref{sec:sensitivity to data-copying}, now including the $C_T$ statistic for comparison. Appendix \ref{sec:appendix moons kDE} provides more experimental details, and examples of the dataset. 

\mypara{Results.} \textbf{Figure \ref{fig:moons gen gap}} depicts how the generalization gap dwindles as KDE $\sigma$ increases. While this test is capable of capturing data-copying, it is insensitive to underfitting and relies on a tractable likelihood. \textbf{Figures \ref{fig:moons z score}} and \textbf{ \ref{fig:moons NN}} give a side-by-side depiction of $C_T$ and the two-sample NN test accuracies across a range of KDE $\sigma$ values. Think of $C_T$ values as $z$-score standard deviations. We see that the $C_T$ statistic in \textbf{Figure \ref{fig:moons z score}} precisely identifies the MLE model when $C_T \approx 0$, and responds sharply to $\sigma$ values above and below $\sigma_{\text{MLE}}$. The baseline in \textbf{Figure \ref{fig:moons NN}} similarly identifies the MLE $Q$ model when training accuracy $\approx 0.5$, but is higher variance and less sensitive to changes in $\sigma$, especially for over-fit $\sigma \ll \sigma_{\text{MLE}}$. We will see in the next experiment, that this test breaks down for more complex datasets when $m \ll |T|$. 

\paragraph{KDEs: MNIST Handwritten Digits.}
\label{sec:MNIST KDE}
We now extend the KDE test performed on the moons dataset to the significantly more complex MNIST handwritten digit dataset \citep{lecun}. 

While it would be convenient to directly apply the KDE $\sigma$-sweeping tests discussed in the previous section, there are two primary barriers. The first is that KDE model relies on $L_2$ norms being perceptually meaningful, which is well understood not to be true in pixel space. The second problem is that of dimensionality: the 784-dimensional space of digits is far too high for a KDE to be even remotely efficient at interpolating the space. 

To handle these issues, we first embed each image, $x \in \calX$, to a perceptually meaningful 64-dimensional latent code, $z \in \mathcal{Z}$. We achieve this by training a convolutional autoencoder with a VGGnet perceptual loss produced by \cite{zhang} (see Appendix \ref{sec:appendix MNIST autoencoder} for more detail). Surely, even in the lower 64-dimensional space, the KDE will suffer some from the curse of dimensionality. We are not promoting this method as a powerful generative model, but rather as an instructive tool for probing a test's response to data-copying in the image domain. All tests are run in the compressed latent space; Appendix \ref{sec:appendix MNIST autoencoder} provides more experimental details. 

As discussesd briefly in Section \ref{sec:sensitivity to data-copying}, a limitation of the two-sample NN test is that it requires $m = |T|$. For a large training set like MNIST, it is computationally challenging to generate $|T|$ samples, even with a 64-dimensional KDE. We therefore use a subsampled training set $\tilde{T}$ of size $10,000 = m < |T|$ when running the two-sample NN test. The proposed $C_T(P_n, Q_m)$ test has no such restriction on the size of $T$ and $Q_m$.

\mypara{Results.} The likelihood generalization gap is depicted in \textbf{Figure \ref{fig:mnist gen gap}} repeating the trend seen with the `moons' dataset. 

\textbf{Figure \ref{fig:mnist kde z score}} shows how $C_T(P_n, Q_m)$ reacts decisively to over- and underfitting. It falsely determines the MLE $\sigma$ value as slightly over-fit. However, the region of where $C_T$ transitions from over- to underfit (say $-13 \leq C_T \leq 13$) is relatively tight and includes the MLE $\sigma$.

Meanwhile, \textbf{Figure \ref{fig:mnist kde NN}} shows how --- with the generated sample smaller than the training sample, $m \ll |T|$ --- the two-sample NN baseline provides no meaningful estimate of data-copying. In fact, the most data-copying models with low $\sigma$ achieve the best scores closest to $0.5$. Again, we are forced to use the $m$-subsampled $\widetilde{T} \subset T$, and most instances of data copying are completely missed.

These results are promising, and demonstrate the reliability of this hypothesis testing approach to probing for data-copying across different data domains. In the next section, we explore how these tests perform on more sophisticated, non-KDE models. 


\subsubsection{Variational Autoencoders}
\label{sec:neural model tests}
Gaussian KDE's may have nice theoretical properties, but are relatively ineffective in high-dimensional settings, precluding domains like images. As such, we also demonstrate our experiments on more practical neural models trained on higher dimensional image datasets (MNIST and ImageNet), with the goal of observing whether the $C_T$ statistic indicates data-copying as these models range from over- to underfit. The first neural model we consider is a Variational Autoencoder (VAE) trained on the MNIST handwritten images dataset. 


\begin{figure*}[h]
    \centering
    \begin{subfigure}{0.49\linewidth}
        \centering
        \includegraphics[width = 1\linewidth]{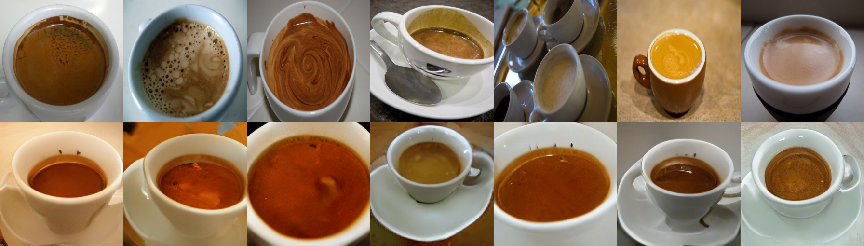}
        \label{fig:biggan coffee overfit}
    \end{subfigure}
    \hfill
    \begin{subfigure}{0.49\linewidth}
        \centering
        \includegraphics[width = 1\linewidth]{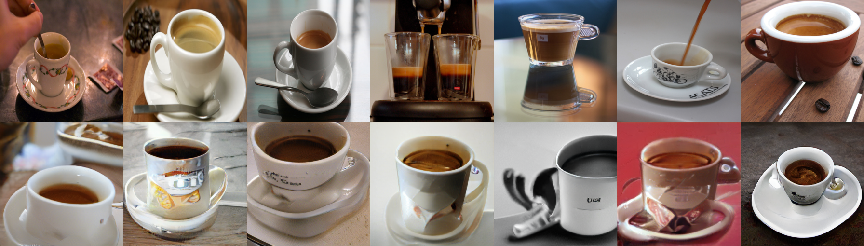}
        \label{fig:biggan coffee underfit}
    \end{subfigure}
    \begin{subfigure}{0.49\linewidth}
        \centering
        \includegraphics[width = 1\linewidth]{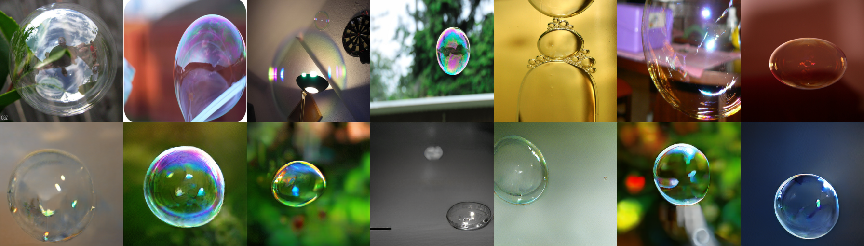}
        \label{fig:biggan bubble overfit}
        \caption{Data-copied cells; top: $Z_U = -1.46$, bottom: $Z_U = -1.00$} 
    \end{subfigure}
    \hfill
    \begin{subfigure}{0.49\linewidth}
        \centering
        \includegraphics[width = 1\linewidth]{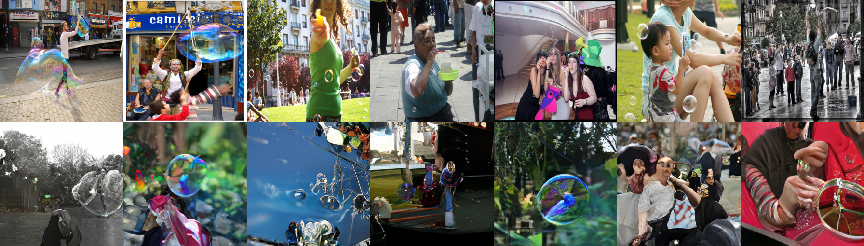}
        \label{fig:biggan bubble underfit}
        \caption{Underfit cells; top: $Z_U = +1.40$, bottom: $Z_U = +0.71$} 
    \end{subfigure}
    \caption{Data-copied and underfit cells of ImageNet12 `coffee' and `soap bubble' instance spaces (trunc. threshold = 2). In each 14-figure strip, the top row provides a random series of training samples from the cell, and the bottom row provides a random series of generated samples from the cell. \textbf{(a)} Data-copied cells. \textbf{(a), top}: Random training and generated samples from a $Z_U = -1.46$ cell of the coffee instance space. \textbf{(a), bottom}: Random training and generated samples from a $Z_U = -1.00$ cell of the bubble instance space. \textbf{(b)} Underfit cells. \textbf{(b), top}: Random training and generated samples from a $Z_U = +1.40$ cell of the coffee instance space. \textbf{(b), bottom}: Random training and generated samples from a $Z_U = +0.71$ cell of the bubble instance space.  }
    \label{fig:biggan overfit underfit}
\end{figure*}


\paragraph{Experimetal Methodology.} Unlike KDEs, VAEs do not have a single parameter that controls the degree of overfitting. Instead, similar to~\cite{VAEs_overfit}, we vary model complexity by increasing the width (neurons per layer) in a three-layer VAE (see Appendix \ref{sec:appendix VAE experiments} for details) -- where higher width means a model of higher complexity. As an embedding, we pass all samples through the the convolutional autoencoder of Section \ref{sec:MNIST KDE}, and collect statistics in this 64-dimensional space. Observe that likelihood is not available for VAEs; instead we compute each model's ELBO on a $10,000$ sample held out validation set, and use the ELBO approximation to the generalization gap instead.

We again note here, that for the NN accuracy baseline, we use a subsampled training set $\tilde{T}$ as with the KDE-based MNIST tests where $m = |\tilde{T}| = 10,000$.  

\paragraph{Results.} \textbf{Figures \ref{fig:vae C_T vs d}} and \textbf{\ref{fig:vae NN vs d}} compare the $C_T$ statistic to the NN accuracy baseline . $C_T$ behaves as it did in the previous sections: more complex models over-fit, forcing $C_T \ll 0$, and less complex models underfit forcing it $\gg0$. We note that the range of $C_T$ values is far less dramatic, which is to be expected since the KDEs were forced to explicitly data-copy. We observe that the ELBO spikes for models with $C_T$ near 0. \textbf{Figure \ref{fig:vae gen gap}} shows the ELBO approximation of the generalization gap as the latent dimension (and number of units in each layer) is decreased. This method is entirely insensitive to over- and underfit models. This may be because the ELBO is only a lower bound, not the actual likelihood. 

The NN baseline in \textbf{Figure \ref{fig:vae NN vs d}} is less interpretable, and fails to capture the overfitting trend as $C_T$ does. While all three test accuracies still follow the upward-sloping trend of \textbf{Figure \ref{fig:moons NN}}, they do not indicate where the highest validation set ELBO is. Furthermore, the NN accuracy statistics are shifted upward when compared to the results of the previous section: all NN accuracies are above 0.5 for all latent dimensions. This is problematic. A test statistic's absolute score ought to bear significance between very different data and model domains like KDEs and VAEs.

\subsubsection{ImageNet GAN}
\label{sec:gan}

Finally, we scale our experiments up to a larger image domain.

\paragraph{Experimental Methodology.} We gather our test statistics on a state of the art conditional GAN, `\emph{BigGan}' \citep{BigGan}, trained on the Imagenet 12 dataset \citep{imagenet12}. Conditioning on an input code, this GAN will generate one of 1000 different Imagenet classes. We run our experiments separately on three classes: `coffee', `soap bubble', and `schooner'. All generated, test, and training images are embedded to a 64-dimensional space by first gathering the 2048-dimensional features of an InceptionV3 network `Pool3' layer, and then projecting them onto the 64 principal components of the training embeddings. Appendix \ref{sec:appendix biggan experiments} has more details. 

Being limited to one pre-trained model, we increase model variance (`truncation threshold') instead of decreasing model complexity. As proposed by \emph{BigGan}'s authors, all standard normal input samples outside of this truncation threshold are resampled. The authors suggest that lower truncation thresholds, by only producing samples at the mode of the input, output higher quality samples at the cost of variety, as determined by Inception Score (IS). Similarly, the FID score finds suitable variety until truncation approaches zero. 

\paragraph{Results.} The results for the $C_T$ score is depicted in \textbf{ Figure \ref{fig:biggan C_T score}}; the statistic remains well below zero until the truncation threshold is nearly maximized, indicating that $Q$ produces samples closer to the training set than real samples tend to be. While FID finds that \emph{in aggregate} the distributions are roughly similar, a closer look suggests that $Q$ allocates too much probability mass near the training samples. 

Meanwhile, the two-sample NN baseline in \textbf{ Figure \ref{fig:biggan NN} } hardly reacts to changes in truncation, even though the generated and training sets are the same size, $m = |T|$. Across all truncation values, the training sample NN accuracy remains around 0.5, not quite implying over- or underfitting.  

A useful feature of the $C_T$ statistic is that one can examine the $Z_U$ scores it is composed of to see which of the cells $\pi \in \Pi_{\tau}$ are or are not copying. \textbf{ Figure \ref{fig:biggan overfit underfit} } shows the samples of over- and underfit clusters for two of the three classes. For both `coffee' and `bubble' classes, the underfit cells are more diverse than the data-copied cells. While it might seem reasonable that these generated samples are further from nearest neighbors in more diverse clusters, keep in mind that the $Z_U > 0$ statistic indicates that they are further from training neighbors than test set samples are. For instance, the people depicted in underfit `bubbles' cell are highly distorted.  

\subsection{Discussion}
We now reflect on the two questions recited at the beginning of Section \ref{sec:experiments}. Firstly, it appears that many existing generative model tests do not detect data-copying. The findings of Section \ref{sec:sensitivity to data-copying} demonstrate that many popular generative model tests like FID, Precision and Recall, and Binning-Based Evaluation are wholly insensitive to explicit data-copying even in low-dimensional settings. We suggest that this is because these tests are geared to detect over- and underrepresentation more than data-copying.

Secondly, the experiments of Section \ref{sec:degree of data-copying} indicate that the proposed $C_T(P_n, Q_m)$ test statistic not only detects explicitly forced data-copying (as in the KDE experiments), but also detects data-copying in complex, overfit generative models like VAEs and GANs. In these settings, we observe that as models overfit more, $C_T$ drops below 0 and significantly below -1. 

A limitation of the proposed $C_T$ test is the number of test samples $n$. Without sufficient test samples, not only is the statistic higher variance, but the instance space partition $\Pi$ cannot be very fine-grain. Consequently, we are limited in our ability to manage the heterogeneity of $d(X)$ across the instance space, and some cells may be mischaracterized. For example, in the BigGan experiment of Section \ref{sec:gan}, we are provided only 50 test samples per image class (e.g. `soap bubble'), limiting us to an instance space partition of only three cells. Developing more data-efficient methods to handle heterogeneity may be a promising area of future work.

\section{Conclusion}
In this work, we have formalized \emph{data-copying}: an under-explored failure mode of generative model overfitting. We have provided preliminary tests for measuring data-copying and experiments indicating its presence in a broad class of generative models. In future work, we plan to establish more theoretical properties of data-copying, convergence guarantees of these tests, and experiments with different model parameters. 

\section{Acknowledgements}
We thank Rich Zemel for pointing us to \cite{Kilian}, which was the starting point of this work. Thanks to Arthur Gretton and Ruslan Salakhutdinov for pointers to prior work, and Philip Isola and Christian Szegedy for helpful advice. Finally, KC and CM would like to thank  ONR under N00014-16-1-2614,  UC Lab Fees under LFR 18-548554 and NSF IIS 1617157 for research support.

\medskip

\bibliographystyle{plainnat}
\bibliography{refs}
\clearpage

\section{Appendix}

\subsection{Proof of Theorem \ref{thm:consistent}}
\label{sec:proof consistent}
A restatement of the theorem: 

\textit{
For true distribution $P$, model distribution $Q$, and distance metric $d:\calX \rightarrow \R$, the estimator $\frac{1}{mn} U_{Q_m} \rightarrow_P \Delta(P,Q)$ according to the concentration inequality 
\begin{align*}
    \pr \big( \ \big|\frac{1}{mn} U_{Q_m} - \Delta(P,Q) \big| \geq t\big) \leq 
    \exp \bigg( -\frac{2 t^2 mn}{m + n} \bigg)
\end{align*}
}
\begin{proof}
We establish consistency using the following nifty lemma 
\begin{lemma}{(Bounded Differences Inequality)}
\label{lem:bounded diff}
Suppose $X_1, \dots, X_n \in \calX$	are independent, and $f:\calX^n \rightarrow \R$. Let $c_1, \dots, c_n$ satisfy 
\begin{align*}
	\sup_{x_1, \dots, x_n, x_i'} \big| &f(x_1, \dots, x_i, \dots, x_n) - f(x_1, \dots, x_i', \dots, x_n) \big| \\
	&\leq c_i
\end{align*}
for $i = 1, \dots, n$. Then we have for any $t > 0$
\begin{align}
	\Pr\big(\big| f - \E[f]\big| \geq t\big) \leq \exp\bigg( \frac{-2t^2}{\sum_{i=1}^n c_i^2} \bigg) 
	\label{eqn:bounded diff}
\end{align}
\end{lemma}

This directly equips us to prove the Theorem. 

It is relatively straightforward to apply Lemma \ref{lem:bounded diff} to the normalized $\overline{U} = \frac{1}{mn} U_{Q_m}$. First, think of it as a function of $m$ independent samples of $X \sim Q$ and $n$ independent samples of $Y \sim P$, $\overline{U}(X_1, \dots, X_m, Y_1, \dots, Y_n) = \frac{1}{mn} \sum_{ij} \mathds{1}_{d(X_i) > d(Y_j)}$
\begin{align*}
	\overline{U}: (\R^{d})^{mn} \rightarrow \R 
\end{align*}
Let $b_i$ bound the change in $\overline{U}$ after substituting any $X_i$ with $X_i'$, and $c_j$ bound the change in $\overline{U}$ after substituting any $Y_j$ with $Y_j'$. Specifically 
\begin{align*}
	\sup_{x_1, \dots, x_m, y_1, \dots, y_n,\  x_i'} \big| &\overline{U}(x_1, \dots, x_i, \dots, x_m, y_1, \dots, y_n) \\
	- &\overline{U}(x_1, \dots, x_i', \dots, x_m, y_1, \dots, y_n)\big| \\
	&\leq b_i \\
	\sup_{x_1, \dots, x_m, y_1, \dots, y_n,\  y_j'} \big| &\overline{U}(x_1, \dots, x_m, y_1, \dots, y_j, \dots  y_n) \\
	- &\overline{U}(x_1, \dots, x_m, y_1, \dots, y_j', \dots  y_n)\big| \\
	&\leq c_i
\end{align*}
We then know that $b_i = \frac{n}{mn} = \frac{1}{m}$ for all $i$, with equality when $d(x_i') < d(y_j) < d(x_i)$ for all $j \in [n]$. In this case, substituting $x_i$ with $x_i'$ flips $n$ of the indicator comparisons in $\overline{U}$ from 1 to 0, and is then normalized by $mn$. By a similar argument, $c_j = \frac{m}{nm} = \frac{1}{n}$ for all $j$. 

Equipped with $b_i$ and $c_j$, we may simply substitute into Equation \ref{eqn:bounded diff} of the Bounded Differences Inequality, giving us 
\begin{align*}
	\Pr\big(\big|  \overline{U}- \E[\overline{U}]\big| \geq t\big) 
	&=  \Pr\big(\big|  \frac{1}{mn} U_{Q_m} - \Delta(\mu_p, \mu_q)\big| \geq t\big)\\
	&\leq \exp\bigg( \frac{-2t^2}{ \sum_{i=1}^m b_i^2 + \sum_{j=1}^n c_j^2} \bigg) \\
	&= \exp\bigg( \frac{-2t^2}{ \sum_{i=1}^m \frac{1}{m^2} + \sum_{j=1}^n \frac{1}{n^2} } \bigg) \\
	&= \exp\bigg( \frac{-2t^2}{ \frac{1}{m} + \frac{1}{n} } \bigg) 
	= \exp\bigg( \frac{-2t^2mn}{ m+n } \bigg)
\end{align*}
\end{proof}

\subsection{Proof of Theorem \ref{thm:fallback}}
\label{sec:proof fallback}
A restatement of the theorem: 

\textit{
When $Q = P$, and the corresponding distance distribution $L(Q) = L(P)$ is non-atomic, 
    \begin{align*}
         \E \Big[ \frac{1}{mn}\overline{U} \Big] &= \frac{1}{2}
    \end{align*}
}
\begin{proof}
	For random variables $A \sim L(P)$ and $B \sim L(P)$, we can partition the event space of $A \times B$ into three disjoint events: 
	\begin{align*}
		\Pr(A > B) &+ \Pr(A < B) \\
		&+ \Pr(A = B) = 1
	\end{align*}
	Since $Q = P$, the first two events have equal probability, $\Pr(A > B) = \Pr(A < B)$, so 
	\begin{align*}
		2\Pr(A > B) + \Pr(A = B) = 1
	\end{align*}
	And since the distributions of $A$ and $B$ are non-atomic (i.e. $\Pr\big(B = b\big) = 0, \quad \forall \ b \in \R$) we have that $\Pr(A = B) = 0$, and thus 
	\begin{align*}
		2\Pr(A > B) &= 1 \\
		\Pr(A > B) &= \Delta(P, Q) =  \frac{1}{2}
	\end{align*}
\end{proof}

\subsection{Proof of Lemma \ref{lemma:kde}} 
\label{sec:appendix kde lemma}
\textbf{Lemma \ref{lemma:kde}}
\textit{
For the kernel density estimator (\ref{eq:kde}), the maximum-likehood choice of $\sigma$, namely the maximizer of $\E_{X \sim P}[\log q_\sigma(X)]$, satisfies
\begin{align*}
	\E_{X \sim P} \bigg[ \sum_{t \in T} Q_\sigma(t|X) &\|X - t\|^2 \bigg] = \\
	&\E_{Y \sim Q_\sigma} \bigg[ \sum_{t \in T} Q_\sigma(t|Y) \|Y - t\|^2 \bigg]
\end{align*}
}
\begin{proof}
We have
\begin{align*}
&\E_{X \sim P} \left[ \ln q_\sigma(X) \right]  \\
&= \E_{X \sim P} \left[ - \ln ((2\pi)^{k/2}|T| \sigma^k) + \ln \sum_{t \in T} \exp\left( -\frac{\|x-t\|^2}{2 \sigma^2}\right)  \right] \\
&=
\mbox{constant} - k \ln \sigma + \E_{X \sim P} \left[\ln \sum_{t \in T} \exp\left( -\frac{\|x-t\|^2}{2 \sigma^2}\right) \right] 
\end{align*}
Setting the derivative of this to zero and simplifying, we find that the maximum-likelihood $\sigma$ satisfies
\begin{equation}
\sigma^2 = \frac{1}{k} \, \E_{X \sim P} \left[ \sum_{t \in T} Q_\sigma(t|X) \|X - t\|^2 \right] .
\label{eq:ml-choice}
\end{equation}
Now, interpreting $Q_\sigma$ as a mixture of $|T|$ Gaussians, and using the notation $t \in_R T$ to mean that $t$ is chosen uniformly at random from $T$, we have
\begin{align*}
	&\E_{Y \sim Q_\sigma} \left[ \sum_{t \in T} Q_\sigma(t|Y) \|Y - t\|^2 \right] \\
	&=\E_{t \in_R T} \E_{Y \sim N(t, \sigma^2 I_k)} \left[ \|Y - t\|^2 \right] 
	= k \sigma^2 .
\end{align*}
Combining this with (\ref{eq:ml-choice}) yields the lemma.
\end{proof}

\subsection{Procedural Details of Experiments}
\subsubsection{Moons Dataset, and Gaussian KDE}
\label{sec:appendix moons kDE}
\begin{figure*}[h]
    \centering
    \begin{subfigure}{.245\linewidth}
        \centering
        \includegraphics[width = 1\linewidth]{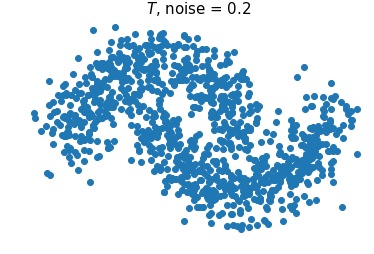}
        \caption{}\label{fig:moons T}
    \end{subfigure}
        \hfill
    \begin{subfigure}{.245\linewidth}
        \centering
        \includegraphics[width = 1\linewidth]{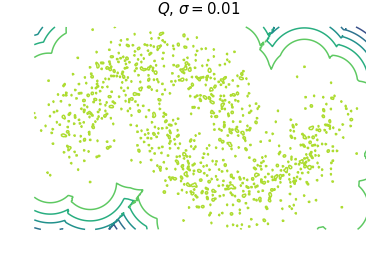}
        \caption{}\label{fig:moons Q .01}
    \end{subfigure}
        \hfill
    \begin{subfigure}{.245\linewidth}
        \centering
        \includegraphics[width = 1\linewidth]{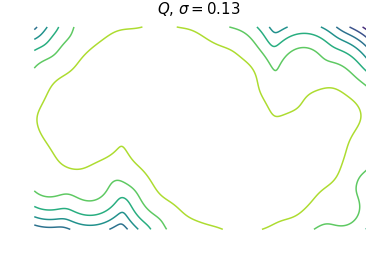}
        \caption{}\label{fig:moons Q .13}
    \end{subfigure}
       \hfill
    \begin{subfigure}{.245\linewidth}
        \centering
        \includegraphics[width = 1\linewidth]{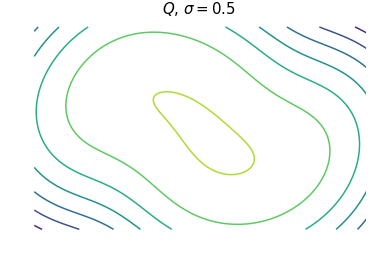}
        \caption{}\label{fig:moons Q .5}
    \end{subfigure}
    \caption{Contour plots of KDE fit on $T$: a) training $T$ sample, b) over-fit `data copying' KDE, c) max likelihood KDE, d) underfit KDE}
    \label{fig:half moon}
\end{figure*} 

\paragraph{moons dataset}
`Moons' is a synthetic dataset consisting of two curved interlocking manifolds with added configurable noise. We chose to use this dataset as a proof of concept because it is low dimensional, and thus KDE friendly and easy to visualize, and we may have unlimited train, test, and validation samples. 

\paragraph{Gaussian KDE}
We use a Gaussian KDE as our preliminary generative model $Q$ because its likelihood is theoretically related to our non-parametric test. Perhaps more importantly, it is trivial to control the degree of data-copying with the bandwidth parameter $\sigma$. $\textbf{Figures \ref{fig:moons Q .01}, \ref{fig:moons Q .13}, \ref{fig:moons Q .5}}$ provide contour plots of of a Gaussian KDE $Q$ trained on the moons dataset with progressively larger $\sigma$. With $\sigma = 0.01$, $Q$ will effectively resample the training set. $\sigma = 0.13$ is nearly the MLE model. With $\sigma = 0.5$, the KDE struggles to capture the unique definition of $T$. 

\subsubsection{Moons Experiments}
\label{sec:appendix moons experiments}
Our experiments that examined whether several baseline tests could detect data-copying (Section \ref{sec:sensitivity to data-copying}), and our first test of our own metric (Section \ref{sec:KDE experiments}) use the moons dataset. In both of these, we fix a training sample, $T$ of 2000 points, a test sample $P_n$ of 1000 points, and a generated sample $Q_m$ of 1000 points. We regenerate $Q_m$ 10 times, and report the average statistic across these trials along with a single standard deviation. If the standard deviation buffer along the line is not visible, it is because the standard deviation is relatively small. We artificially set the constraint that $m,n \ll |T|$, as is true for big natural datasets, and more elaborate models that are computationally burdensome to sample from. 

\paragraph{Section \ref{sec:sensitivity to data-copying} Methods} Here are the routines we used for the four baseline tests: 
\begin{itemize}
    \item \textbf{Frech\'et Inception Distance (FID)} \citep{heusel}: Normally, this test is run on two samples of images ($T$ and $Q_m$) that are first embedded into a perceptually meaningful latent space using a discriminative neural net, like the Inception Network. By `meaningful' we mean points that are closer together are more perceptually alike to the human eye. Unlike images in pixel space, the samples of the moons dataset require no embedding, so we run the Frech\'et test directly on the samples. 
    
    First, we fit two MLE Gaussians: $\mathcal{N}(\mu_T, \Sigma_T)$ to $T$, and $\mathcal{N}(\mu_Q, \Sigma_Q)$ to $Q_m$, by collecting their respective MLE mean and covariance parameters. The statistic reported is the Frech\'et distance between these two Gaussians, denoted $\text{Fr}(\bullet, \bullet)$, which for Gaussians has a closed form: 
    \begin{align*}
        \text{Fr}&\big(\mathcal{N}(\mu_T, \Sigma_T), \mathcal{N}(\mu_Q, \Sigma_Q)\big) = \\
         &\|\mu_T - \mu_Q\| + \textbf{Tr}\big( \Sigma_T - \Sigma_Q - 2(\Sigma_T \Sigma_Q)^{\nicefrac{1}{2}} \big)
    \end{align*}
    Naturally, if $Q$ is data-copying $T$, its MLE mean and covariance will be nearly identical, rendering this test ineffective for capturing this kind of overfitting. 
    
    \item \textbf{Binning Based Evaluation} \citep{richardson}: This test, takes a hypothesis testing approach for evaluating mode collapse and deletion. The test bears much similarity to the test described in Section \ref{sec:local-versus-global}. The basic idea is as follows. Split the training set into partition $\Pi$ using $k$-means; the number of samples falling into each bin is approximately normally distributed if it has >20 samples. Check the null hypothesis that the normal distribution of the fraction of the training set in bin $\pi$, $T(\pi)$, equals the normal distribution of the fraction of the generated set in bin $\pi$, $Q_m(\pi)$. Specifically: 
    \begin{align*}
        Z_\pi = \frac{Q_m(\pi) - T(\pi)}
         {\sqrt{ \widehat{p}\big(1 - \widehat{p}\big) \big( \frac{1}{|T|} + \frac{1}{m} \big) }}
    \end{align*}
    where $\widehat{p} = \frac{|T|T(\pi) + mQ_m(\pi)}{|T| + m}$. We then perform a one-sided hypothesis test, and compute the number of positive $Z_\pi$ values that are greater than the significance level of 0.05. We call this the number of statistically different bins or NDB. The NDB/$k$ ought to equal the significance level if $P = Q$. 
    \item \textbf{Two-Sample Nearest-Neighbor} \citep{lopez}: In this test --- our primary baseline --- we report the three LOO NN values discussed in \cite{Kilian}. The generated sample $Q_m$ and training sample (subsampled to have equal size, $m$), $\widetilde{T} \subseteq T$, are joined together create sample $S = \widetilde{T} \cup Q_m$ of size $2m$, with training samples labeled `1' and test samples labeled `0'. One then fits a 1-Nearest-Neighbor classifier to $S$, and reports the accuracy in predicting the training samples (`1's), the accuracy in predicting the generated samples (`0's), and the average. 
    
    One can expect that --- when $Q$ collapses to a few mode centers of $T$ --- the training accuracy is low, and the generated accuracy is high, thus indicating over-representation. Additionally, one could imagine that when the training and generated accuracies are near 0, we have extreme data-copying. However, as explained in Experiments section, when we are forced to subsample $T$, it is unlikely that a given copied training point $t \in T$ is used in the test, thus making the test result unclear. 
    \item \textbf{Precision and Recall} \citep{mehdi}: This method offers a clever technique for scaling classical precision and recall statistics to high dimensional, complex spaces. First, all samples are embedded to Inception Network Pool3 features. Then, the author's use the following insight: for distribution's $Q$ and $P$, the precision and recall curve is approximately given by the set of points: 
    \begin{align*}
        \widehat{\text{PRD}}(Q,P) &= \{(\alpha(\lambda), \beta(\lambda) | \lambda \in \Lambda \}
    \end{align*}
    where
    \begin{align*}
        \Lambda &= \{ \tan\big( \frac{i}{r + 1} \frac{\pi}{2} \big) | i \in [r] \} \\
        \alpha(\lambda) &= \sum_{\pi \in \Pi} \min \big( \lambda P(\pi), Q(\pi) \big) \\
        \beta(\lambda) &= \sum_{\pi \in \Pi} \min \big( P(\pi), \frac{Q(\pi)}{\lambda} \big) 
    \end{align*}
    and where $r$ is the `resolution' of the curve, the set $\Pi$ is a partition of the instance space and $P(\pi), Q(\pi)$ are the fraction of samples falling in cell $\pi$. $\Pi$ is determined by running $k$-means on the combination of the training and generated sets. In our tests here, we set $k = 5$, and report the average PRD curve measured over 10 $k$-means clusterings (and then re-run 10 times for 10 separate trials of $Q_m$). 
    
    \end{itemize}

\subsubsection{MNIST Experiments}
\label{sec:appendix MNIST autoencoder}
\begin{figure*}[h]
	\centering
	\includegraphics[width = \linewidth]{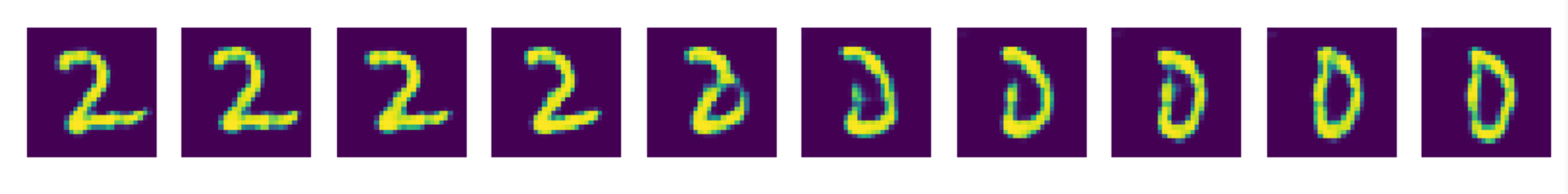}
	\caption{Interpolating between two points in the latent space to demonstrate $L_2$ perceptual significance}
	\label{fig:latent_interp}
\end{figure*}

\begin{figure*}[h]
    \centering
    \begin{subfigure}{.31\linewidth}
        \centering
        \includegraphics[width = 1\linewidth]{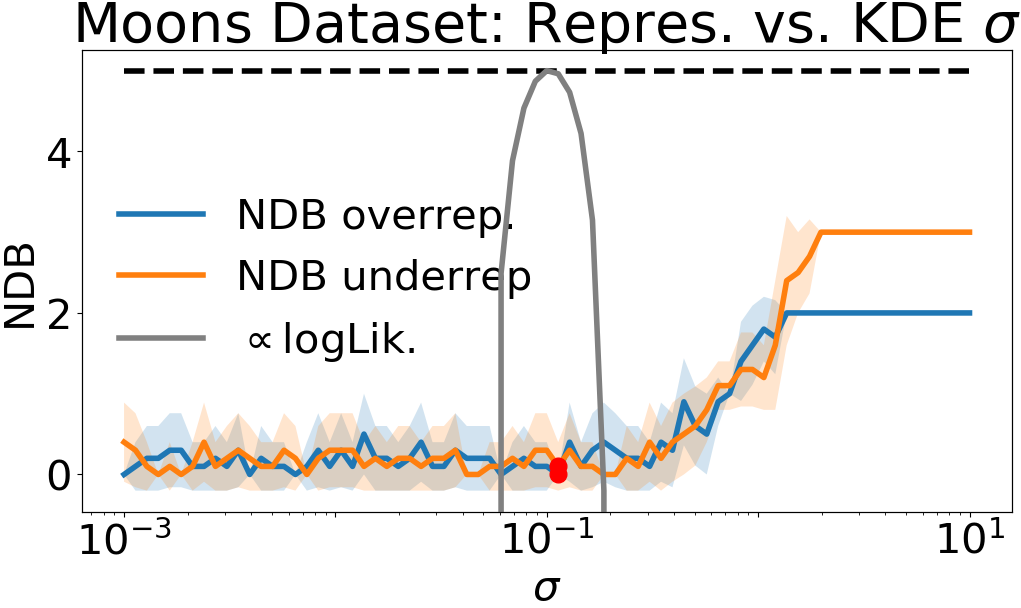}
        \caption{}\label{fig:moons KDE rep}
    \end{subfigure}
        \hfill
    \begin{subfigure}{.31\linewidth}
        \centering
        \includegraphics[width = 1\linewidth]{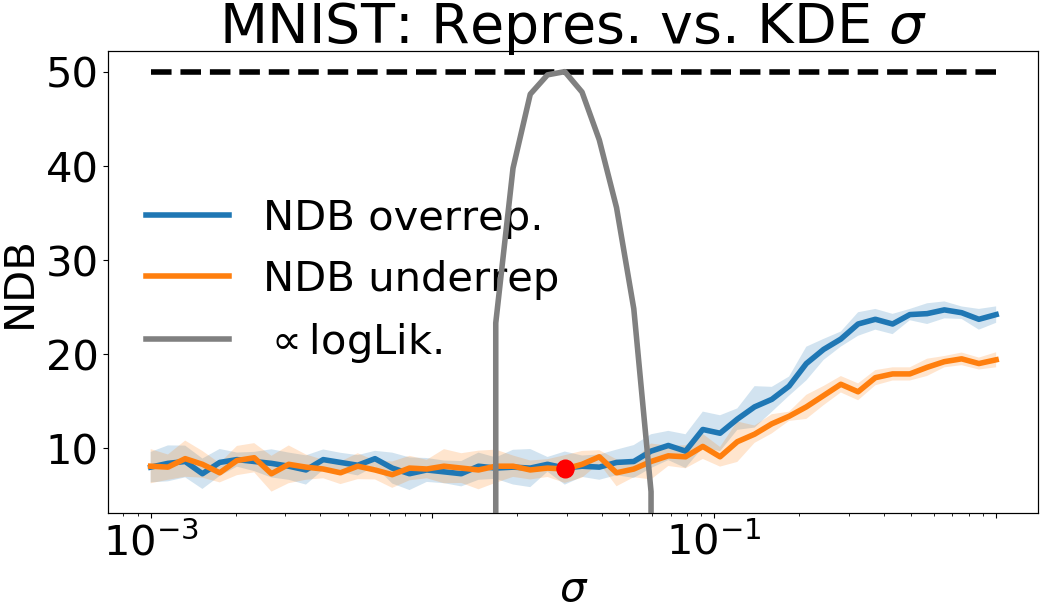}
        \caption{}\label{fig:mnist kde rep}
    \end{subfigure}
        \hfill
    \begin{subfigure}{.31\linewidth}
        \centering
        \includegraphics[width = 1\linewidth]{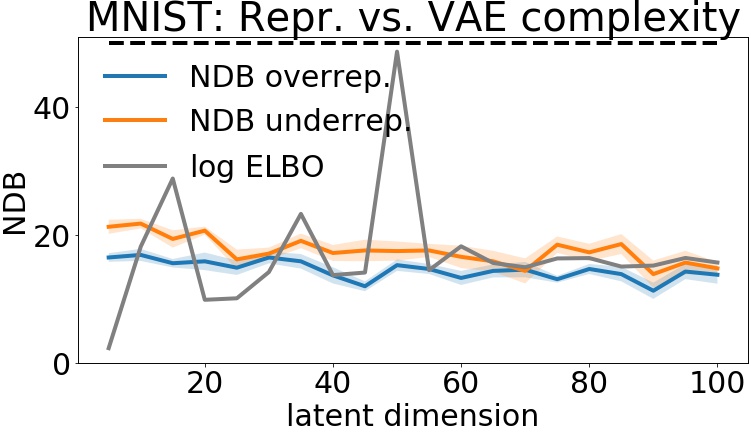}
        \caption{}\label{fig:vae rep}
    \end{subfigure}
    \caption{Number of statistically different bins, both those over and under the significance level of 0.05. The black dotted line indicates the total number of cells or `bins'. \textbf{(a,b)} KDEs tend to start misrepresenting with $\sigma \gg \sigma_{\text{MLE}}$, which makes sense as they become less and less dependent on training set. \textbf{(c)} it makes sense that the VAE over- and under-represents across all latent dimensions due to its reverse KL loss. There is slightly worse over- and under-representation for simple models with low latent dimension. }
    \label{fig:rep tests 1}
\end{figure*} 

\begin{figure*}[h]
    \centering
    \begin{subfigure}{.31\linewidth}
        \centering
        \includegraphics[width = 1\linewidth]{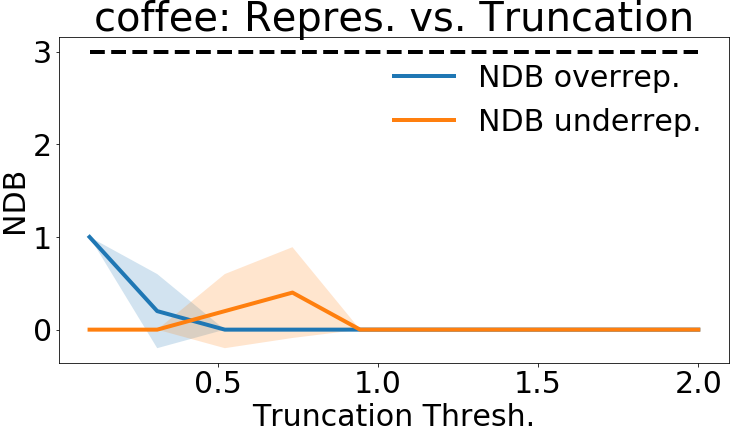}
        \caption{}\label{fig:moons KDE rep}
    \end{subfigure}
        \hfill
    \begin{subfigure}{.31\linewidth}
        \centering
        \includegraphics[width = 1\linewidth]{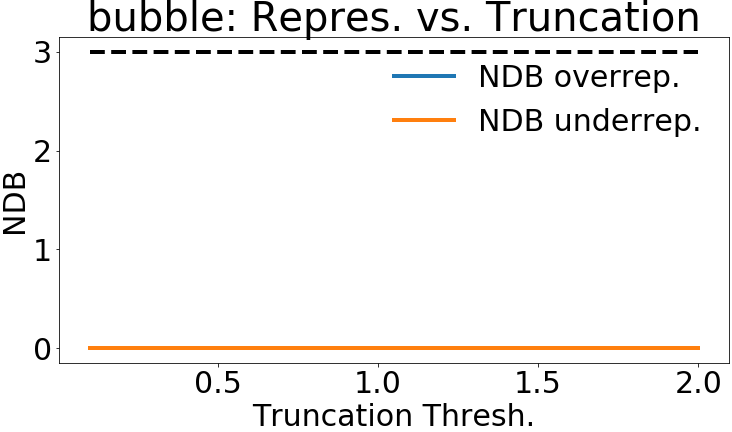}
        \caption{}\label{fig:mnist kde rep}
    \end{subfigure}
        \hfill
    \begin{subfigure}{.31\linewidth}
        \centering
        \includegraphics[width = 1\linewidth]{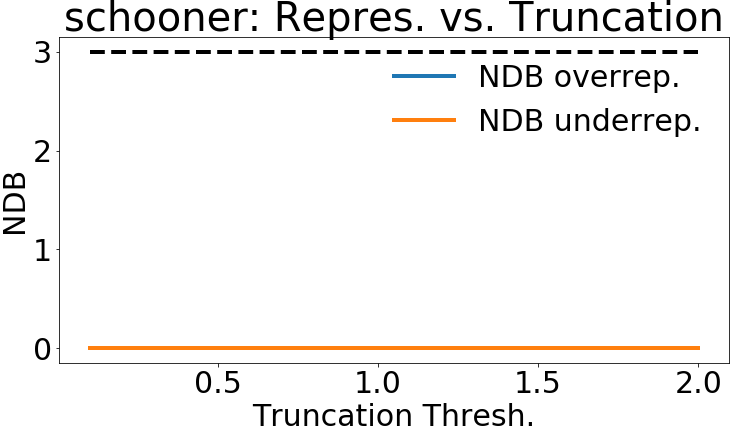}
        \caption{}\label{fig:vae rep}
    \end{subfigure}
    \caption{This GAN model produces relatively equal representation according to our clustering for all three classes. It makes sense that a low truncation level tends to over-represent for one class, as a lower truncation threshold causes less variance. Even though it places samples into all cells, some cells are data-copying much more aggressively than others.}
    \label{fig:biggan rep tests }
\end{figure*} 
The experiments of Sections \ref{sec:MNIST KDE} and \ref{sec:neural model tests} use the MNIST digit dataset \citep{lecun}. We use a training sample, $T$, of size $|T| = 50,000$, a test sample $P_n$ of size $n = 10,000$, a validation sample $V_l$ of $l = 10,000$, and create generated samples of size $m = 10,000$. 

Here, for a meaningful distance metric, we create a custom embedding using a convolutional autoencoder trained using a VGG perceptual loss proposed by \cite{zhang}. The encoder and decoder each have four convolutional layers using batch normalization, two linear layers using dropout, and two max pool layers. The autoencoder is trained for 100 epochs with a batch size of 128 and Adam optimizer with learning rate 0.001. For each training sample $t \in \R^{784}$, the encoder compresses to $z \in \R^{64}$, and decoder expands back up to $\widehat{t} \in \R^{784}$. Our loss is then 
\begin{align*}
    L(t, \widehat{t}) = \gamma(t, \widehat{t}) +  \lambda  \max \{\| z \|_2^2 - 1, 0  \}
\end{align*}
where $\gamma(\bullet, \bullet)$ is the VGG perceptual loss, and $ \lambda  \max \{\| z \|_2^2 - 1, 0  \}$ provides a linear hinge loss outside of a unit $L_2$ ball. The hinge loss encourages the encoder to learn a latent representation within a bounded domain, hopefully augmenting its ability to interpolate between samples. It is worth noting that the perceptual loss is not trained on MNIST, and hopefully uses agnostic features that help keep us from overfitting. We opt to use a standard autoencoder instead of a stochastic autoencoder like a VAE, because we want to be able to exactly data-copy the training set $T$. Thus, we want the encoder to create a near-exact encoding and decoding of the training samples specifically. \textbf{Figure \ref{fig:latent_interp}} provides an example of linearly spaced steps between two training samples. While not perfect, we observe that half-way between the `2' and the `0' is a sample that appears perceptually to be almost almost a `2' and almost a `0'. As such, we consider the distance metric $d(x)$ on this space used in our experiments to be meaningful. 

\begin{figure*}[h]
    \centering
    \begin{subfigure}{0.49\linewidth}
        \centering
        \includegraphics[width = 1\linewidth]{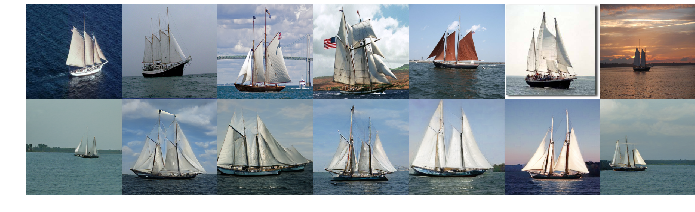}
        \label{fig:biggan schooner overfit}
        \caption{$Z_U = -1.05$}
    \end{subfigure}
    \hfill
    \begin{subfigure}{0.49\linewidth}
        \centering
        \includegraphics[width = 1\linewidth]{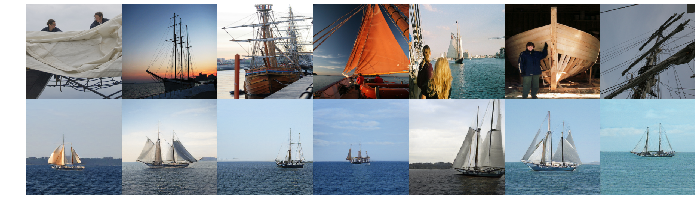}
        \label{fig:biggan schooner underfit}
        \caption{$Z_U = +1.32$}
    \end{subfigure}
    \caption{Example of data-copied and underfit cell of ImageNet `schooner' instance space, from `BigGan' with trunc. threshold = 2. We note here, that --- limited to only 50 training samples --- the insufficient $k = 3$ clustering is perhaps not fine grain enough for this class. Notice that the generated samples falling into the underfit cell (mostly training images of either masts or fronts of boats) are hardly any different from those of the over-fit cell.  They are likely on the boundary of the two cells. With that said, the samples of the data-copied cell \textbf{(a)} are certainly close to the training samples in this region.}
    \label{fig:biggan schooner examples}
\end{figure*}

\paragraph{\textbf{KDE tests}:}
In the MNIST KDE experiments, we fit each KDE $Q$ on the 64-d latent representations of the training set $T$ for several values of $\sigma$; we gather all statistical tests in this space, and effectively only decode to visaully inspect samples. We gather the average and standard deviation of each data point across 5 trials of generating $Q_m$. For the Two-Sample Nearest-Neighbor test, it is computationally intense to compute the nearesnt neighbor in a 64-dimensional dataset of 20,000 points $\widetilde{T} \cup Q_m$ 20,000 times. To limit this, we average each of the training and generated NN accuracy over 500 training and generated samples. We find this acceptable, since the test results depicted in \textbf{Figure \ref{fig:mnist kde NN}} are relatively low variance. 

\paragraph{VAE experiments:}
\label{sec:appendix VAE experiments}
In the MNIST VAE experiments, we only use the 64-d autoencoder latent representation in computing the $C_T$ and 1-NN test scores, and not at all in training. Here, we experiment with twenty standard, fully connected, VAEs using binary cross entropy reconstruction loss. The twenty models have three hidden layers and latent dimensions ranging from $d = 5$ to $d = 100$ in steps of 5. The number of neurons in intermediate layers is approximately twice the number of the layer beneath it, so for a latent space of 50-d, the encoder architecture is $784 \rightarrow 400 \rightarrow 200 \rightarrow 100 \rightarrow 50$, and the decoder architecture is the opposite. 

To sample from a trained VAE, we sample from a standard normal with dimensionality equivalent to the VAEs latent dimension, and pass them through the VAE decoder to the 784-d image space. We then encode these generated images to the agnostic 64-d latent space of the perceptual autoencoder described at the beginning of the section, where $L_2$ distance is meaningful. We also encode the training sample $T$ and test sample $P_n$ to this space, and then run the $C_T$ and two-sample NN tests. We again compute the nearest neighbor accuracies for 10,000 of the training and generated samples (the 1-NN classifier is fit on the 20,000 sample set $\widetilde{T} \cup Q_m$), which appears to be acceptable due to low test variance. 

\subsubsection{ImageNet Experiments}
\label{sec:appendix biggan experiments}
Here, we have chosen three of the one thousand ImageNet12 classes that `BigGan' produces. To reiterate, a conditional GAN can output samples from a specific class by conditioning on a class code input. We acknowledge that conditional GANs combine features from many classes in ways not yet well understood, but treat the GAN of each class as a uniquely different generative model trained on the training samples from that class. So, for the `coffee' class, we treat the GAN as a coffee generator $Q$, trained on the 1300 `coffee' class samples. For each class, we have 1300 training samples $|T|$, 2000 generated samples $m$, and 50 test samples $n$. Being atypically training sample starved ($m > |T|$), we subsample $Q_m$ (not $T$!), to produce equal size samples for the two-sample NN test. As such, all training samples used are in the combined set $S$. We also note that the 50 test samples provided in each class is highly limiting, only allowing us to split the instance space into about three cells and keep a reasonable number of test samples in each cell. As the number of test samples grows, so can the number of cells and the resolution of the partition. \textbf{Figure \ref{fig:biggan schooner examples}} provides an example of where this clustering might be limited; the generated samples of the underfit cell seem hardly any different from those of the over-fit cell. A finer-grain partition is likely needed here. However, the data-copied cell to the left does appear to be very close to the training set, potentially too close according to $Z_U$. 

In performing these experiments, we gather the $C_T(P_n, Q_m)$ statistic for a given class of images. In an attempt to embed the images into a lower dimensional latent space with $L_2$ significance, we pass each image through an InceptionV3 network and gather the 2048-dimension feature embeddings after the final average pooling layer (Pool3). We then project all inception-space images ($T, P_n, Q_m$) onto the 64 principal components of the training set embeddings. Finally, we use $k$-means to partition the points of each sample into one of $k = 3$ cells. The number of cells is limited by the 50 test images available per class. Any more cells would strain the Central Limit Theorem assumption in computing $Z_U$. Finally, we gather the $C_T$ and two-sample NN baseline statistics on this 64-d space. 

\begin{figure*}[h]
    \centering
    \begin{subfigure}{0.32\linewidth}
        \centering
        \includegraphics[width = 1\linewidth]{./images/moons_z_score.png}
        \caption{ }
        \label{fig:kMMD comparison moons c_t}
    \end{subfigure}
    \hfill
    \begin{subfigure}{0.32\linewidth}
        \centering
        \includegraphics[width = 1\linewidth]{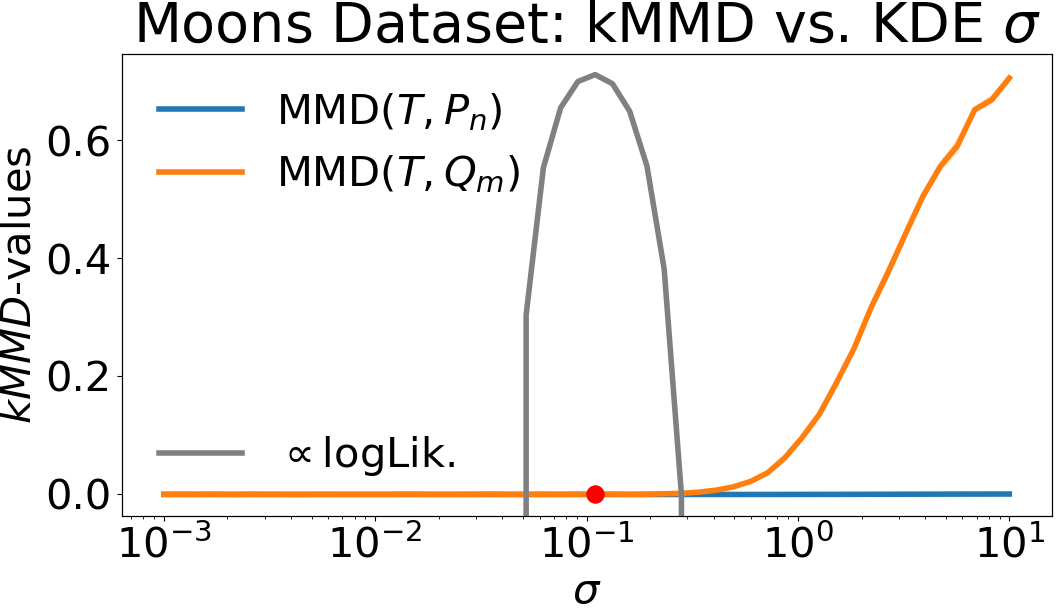}
        \caption{ }
        \label{fig:kMMD comparison moons kMMD}
    \end{subfigure}
    \begin{subfigure}{0.32\linewidth}
        \centering
        \includegraphics[width = 1\linewidth]{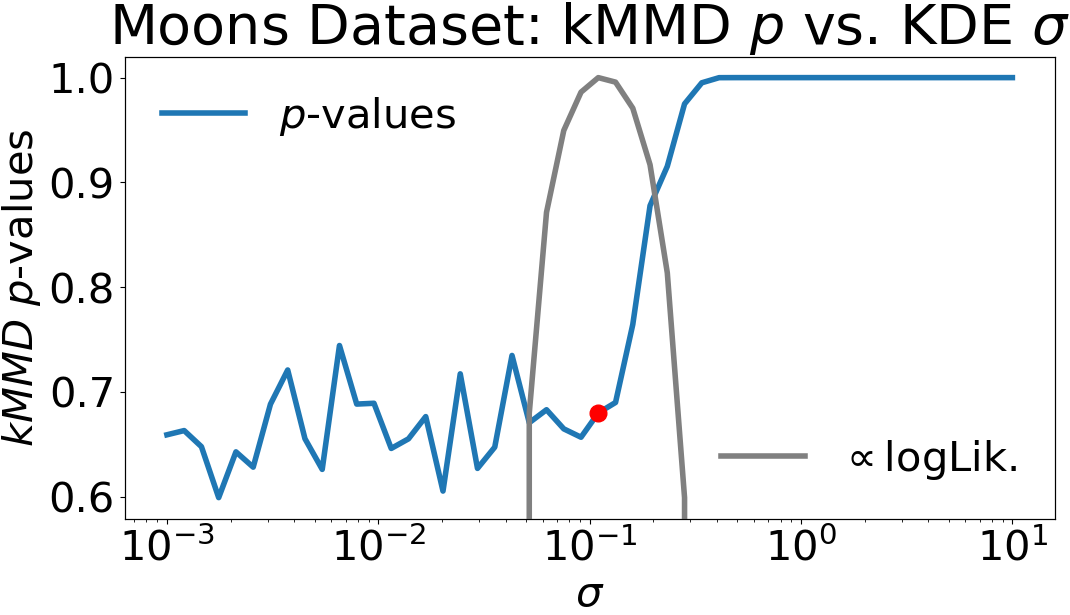}
        \caption{ }
        \label{fig:kMMD comparison moons kMMD p}
    \end{subfigure}
    \begin{subfigure}{0.47\linewidth}
        \centering
        \includegraphics[width = 1\linewidth]{./images/VAE_C_T_vs_d.png}
        \caption{ }
        \label{fig:kMMD comparison vae}
    \end{subfigure}
    \hfill
    \begin{subfigure}{0.49\linewidth}
        \centering
        \includegraphics[width = 1\linewidth]{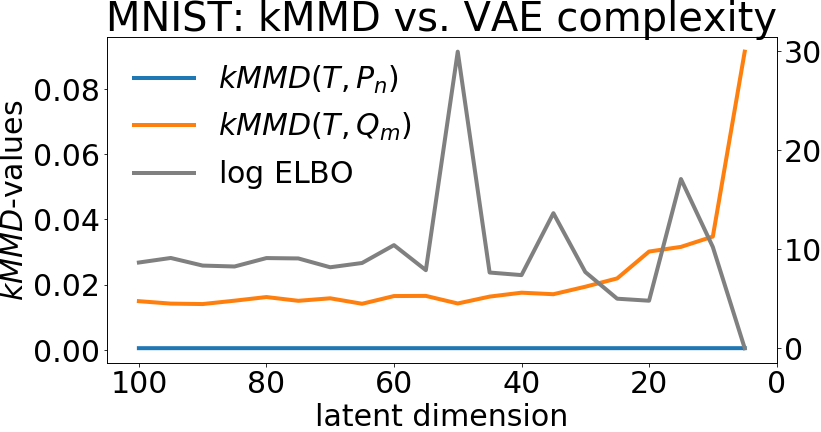}
        \caption{ }
        \label{fig:kMMD comparison kMMD}
    \end{subfigure}
    \caption{Comparison of the $C_T(P_n, Q_m)$ test presented in this paper alongside the three sample kMMD test. \textbf{(a)} and \textbf{(b)} compare the two tests on the simple moons dataset where $Q$ is a gaussian KDE (repeating the experiments of \textbf{Figure \ref{fig:compare methods}}). \textbf{(c)} and \textbf{(d)} compare the two tests for the MNIST dataset where $Q$ is a VAE (repeating the experiments of \textbf{Figure \ref{fig:neural experiments}}. } 
    \label{fig:kMMD comparison}
\end{figure*}

\subsection{Comparison with three-sample Kernel-MMD:}
\label{sec:kMMD}
Another three-sample test not shown in the main body of this work is the three-sample kernel MMD test introduced by \cite{gretton_2} intended more for model comparison than for checking model overfitting. For samples $X \sim P$ and $Y \sim Q$, we can estimate the squared kernel MMD between $P$ and $Q$ under kernel $k$ by empirically estimating 

\begin{align*}
    &\text{MMD}^2(P,Q) \\
    &= \E_{x,x' \sim P}[k(x,x')] - 2\E_{x\sim P, y \sim Q}[k(x,y)] +  \E_{y,y' \sim Q}[k(y,y')]
\end{align*}

More recent works such as \cite{reviewer_paper} have repurposed this test for measuring generative model overfitting. Intuitively, if the model is overfitting its training set, the empirical kMMD between training and generated data may be smaller than that between training and test sets. This may be triggered by the data-copying variety of overfitting. 

This test provides an interesting benchmark to consider in addition to those in the main body. \textbf{Figure \ref{fig:kMMD comparison}} demonstrates some preliminary experimental results repeating both the `moons' KDE experiment of \textbf{Figure \ref{fig:compare methods}} and the MNIST VAE experiment \textbf{Figure \ref{fig:vae C_T vs d}}. To implement the kMMD test, we used code posted by \cite{gretton_2} https://github.com/eugenium/MMD, specifically the three sample RBF-kMMD test. 

In \textbf{Figures \ref{fig:kMMD comparison moons c_t}}, \textbf{\ref{fig:kMMD comparison moons kMMD}}, and \textbf{\ref{fig:kMMD comparison moons kMMD p}} we compare $C_T(P_n, Q_m)$ and the kMMD gap respectively for 50 values of KDE $\sigma$. We observe that the kMMD between the test and training set (blue) and between the generated and training set (orange) remain near zero for all $\sigma$ values less than the MLE $\sigma$, indicated by the red circle. This suggests that the three-sample kMMD is not a particularly strong test for data-copying, since low $\sigma$ values are effectively bootstrapping the original training set. The kMMD gap does diverge for $\sigma$ values much larger than the MLE $\sigma$, indicating that it can detect underfitting by $Q$, however. 

This is corroborated by \textbf{Figure \ref{fig:kMMD comparison moons kMMD p}} which displays the $p$-value of the kMMD hypothesis test used by \cite{reviewer_paper}. This checks the null hypothesis that the kMMD between $T$ and $Q_m$ is greater than that between $T$ and $P_n$. A high $p$-value confirms this null hypothesis (as seen for all $\sigma > \sigma_{\text{MLE}}$). A $p$-value near 0.5 suggests that the kMMD's are approximately equal. A $p$-value near zero rejects the null hypothesis, suggesting that the kMMD between $Q_m$ and $T$ is much smaller than that between $P_n$ and $T$. We see that the $p$-value remains well above 0.5 for all $\sigma$ values and treats the MLE $\sigma$ just as it does the overfit $\sigma$ values.  

\textbf{Figures \ref{fig:kMMD comparison vae}} and \textbf{\ref{fig:kMMD comparison kMMD}} compare the $C_T$ and kMMD tests for twenty MNIST VAEs with decreasing complexity (latent dimension). \textbf{Figures \ref{fig:kMMD comparison kMMD}} again depicts the kMMD distance to training set for both the generated (orange) and test samples (blue). We observe that this test does not appear sensitive to over-parametrized VAEs ($d > 50$) in the same way our proposed test (\textbf{Figure \ref{fig:kMMD comparison vae}}) is. As in the `moons' case above, it appears sensitive to underfitting ($d << 50$). Here, the corresponding kMMD $p$-values are effectively 1 for all latent dimension values, and thus are omitted. 

We suspect that this insensitivity to data-copying is due to the fact that -- for a large number of samples $m,n$ -- the kMMDs between $T$ and $P_n$, and between $T$ and $Q_m$ are both likely to be near zero when $Q$ data-copies. Consider the case of extreme data-copying, when $Q_m$ is simply a bootstrap sample from the training set $T$. The kMMD estimate will be 
\begin{align*}
    &\widehat{\text{MMD}^2}(T,Q_m) \\
    &= \sum_{x,x' \in T}k(x,x') - 2\sum_{x\in T \atop y \in Q_m}k(x,y) +  \sum_{y,y' \in Q_m}k(y,y')
\end{align*}
Informally speaking, the second and third summations of this expression almost behave identically on average to the first since $Q_m$ is a bootstrap sample. They only behave differently for summation terms that are collisions: $k(x,y), x=y$ in the second summation, and $k(y,y'), y=y'$ in the third summation. The fraction of these collision summation terms decreases rapidly with $m,n$. Consequently, the kMMD between the training set and this bootstrap sample is very near zero. It is clear to see that the same goes for the kMMD between the training and test sample, since they are independent samples from the same distribution. 

The experiment in \textbf{Figure \ref{fig:kMMD comparison kMMD}} does not test for extreme data-copying since the model is a VAE, not KDE. Consequently, there still is an appreciable gap between the test and generated kMMD to the training set. Regardless, this kMMD is not sensitive enough to differentiate between the well-fit model of dimension $\approx 50$ and the overfit models of dimension $\approx 100$.  

\end{document}